\documentclass{article}
\usepackage[preprint]{neurips_2019}

\usepackage[utf8]{inputenc} % allow utf-8 input
\usepackage[T1]{fontenc}    % use 8-bit T1 fonts
\usepackage{hyperref}       % hyperlinks
\usepackage{url}            % simple URL typesetting
\usepackage{booktabs}       % professional-quality tables
\usepackage{amsfonts}       % blackboard math symbols
\usepackage{nicefrac}       % compact symbols for 1/2, etc.
\usepackage{microtype}      % microtypography
\usepackage{wrapfig}

\usepackage[noend]{algpseudocode}
\algrenewcommand{\algorithmiccomment}[1]{\hfill (#1)}
\usepackage{xpatch}

\algrenewcommand\alglinenumber[1]{{\footnotesize#1}}

\makeatletter
\xpatchcmd{\algorithmic}{\itemsep\z@}{\itemsep=0.5ex plus2pt}{}{}
\makeatother

\usepackage{algorithm}
\usepackage{graphicx}
\usepackage{subcaption}
\usepackage{amsmath, amssymb, amsthm}
\usepackage{physics}
\usepackage{hyperref}
\usepackage{color, xcolor, pagecolor}
\usepackage{xparse, booktabs}
\usepackage{enumitem}

\usepackage{comment}

\newcommand{\R}{\mathbb{R}} % Real Numbers
\newcommand{\E}{\mathbb{E}} % Expectation
\newcommand{\lossf}{l} % Loss function
\newcommand{\Lagr}{\mathcal{L}} % Lagrangian
\newcommand{\D}{D} % Bregman divergence
\newcommand{\Dom}{\mathcal{D}} % Domain
\newcommand{\prior}{z} % prior variable
\newcommand{\Proj}{\text{P}} % projection
\renewcommand{\l}{\left}
\renewcommand{\r}{\right}
\newcommand{\wh}[1]{#1^*}
\renewcommand{\b}{\beta} % Penalty weight
\renewcommand{\d}{\delta} % Error
\renewcommand{\k}{\kappa} % Convergence rates
\newcommand{\s}{\sigma} % Dual step size
\newcommand{\g}{\gamma} % Primal step size
\newcommand{\ol}[1]{\overline{#1}}
\renewcommand{\t}{\theta}
\renewcommand{\a}{\alpha}
\newcommand{\n}{\natural}

\newtheorem{theorem}{Theorem} % reset theorem numbering for each section

\newtheorem{lemma}[theorem]{Lemma}
\newtheorem{definition}{Definition}
\newtheorem{proposition}{Proposition}
\newtheorem{assumption}{Assumption}

\newcommand{\comparisonfolder}{figures/comparison}

\ExplSyntaxOn
\NewDocumentEnvironment{places}{mm}
 {% #1 is the desired width, #2 is the number of photos per line
  \setlength{\tabcolsep}{0pt} % no space between rows
  \dim_set:Nn \l_places_width_dim
   {
    (#1-\ht\strutbox-\dp\strutbox-2pt)/(#2)
   }
  \begin{tabular}{r @{\hspace{2pt}} *{#2}{c}}
 }
 {
  \end{tabular}
 }

\NewDocumentCommand{\place}{mm}
 {% #1 is the name of the place, #2 is the comma separated list of images
  \seq_set_from_clist:Nn \l_places_images_in_seq { #2 }
  \seq_set_map:NNn \l_places_images_out_seq \l_places_images_in_seq { \places_set_image:n {##1} }
  \seq_put_left:Nn \l_places_images_out_seq
   {
    \begin{tabular}{c}\rotatebox[origin=c]{30}{\strut#1}\end{tabular}
   }
  \seq_use:Nn \l_places_images_out_seq { & } \\ \addlinespace
 }

\dim_new:N \l_places_width_dim
\seq_new:N \l_places_images_in_seq
\seq_new:N \l_places_images_out_seq

\cs_new_protected:Nn \places_set_image:n
 {
  \makebox[\l_places_width_dim]
   {
    \begin{tabular}{c}
    \includegraphics[
      width=\l_places_width_dim,
      height=\l_places_width_dim,
      keepaspectratio,
    ]{#1}
    \end{tabular}
   }
 }

\ExplSyntaxOff

\usepackage{marginnote}

\title{
Fast and Provable ADMM for \\ Learning with Generative Priors
}
\author{%
    Fabian Latorre G\'omez \\EPFL \\ fabian.latorregomez@epfl.ch \And Armin Eftekhari \\ EPFL
    \\ armin.eftekhari@epfl.ch \And Volkan Cevher \\
    EPFL \\ volkan.cevher@epfl.ch
}

\begin{document}
\maketitle

\begin{abstract}
%By leveraging the representation power of modern neural networks,
%Generative Adversarial Networks (GANs) can accurately model data manifolds
%as parametric functions. However, the potential of GANs in modeling has not been fully achieved in the absence a versatile and provable algorithm for solving the resulting non-convex programs. 
In this work, we propose a (linearized) Alternating Direction
Method-of-Multipliers (ADMM) algorithm for minimizing a convex function
subject to a nonconvex constraint. We focus on the special case where such
constraint arises from the specification that a variable should lie in the
range of a neural network. This is motivated by recent successful
applications of Generative Adversarial Networks (GANs) in tasks like
compressive sensing, denoising and robustness against adversarial examples.
The derived rates for our algorithm are characterized in terms of
certain geometric properties of the generator network, which we show hold for
feedforward architectures, under mild assumptions. Unlike gradient descent
(GD), it can efficiently handle non-smooth objectives as well as exploit
efficient partial minimization procedures, thus being faster in many
practical scenarios.
% minimization convex composite non-smooth
%functions over the data manifold encoded by GANs.
\end{abstract}

\section{Introduction}
\label{sec:introduction}
Generative Adversarial Networks (GANs) \citep{Goodfellow2014} show great promise for faithfully modeling complex  data distributions, 
such as natural images \citep{Radford2015,brock2018large} or audio signals
\citep{engel2018gansynth,donahue2019wavegan}. Understanding and improving the theoretical and
practical aspects of their training has thus attracted significant interest~\citep{Lucic2018,Mescheder2018,daskalakis2018training,Hsieh2018,Gidel2019}.
%More recently, the question of how to leverage the modeling power of GANs to improve
%or devise new applications has seen positive and encouraging answers in tasks

Researchers have also begun to leverage the modeling power of GANs  in applications ranging from compressive sensing \citep{Bora2017}, to image denoising
\citep{Lipton2017,Tripathi2018}, to robustness against adversarial examples
\citep{Ilyas2017,Samangouei2018}.

These and other \citep{Dhar2018,Ulyanov2018}  applications  model high-dimensional data as the output of the generator network associated with a GAN, and often lead to a highly non-convex optimization problem of the form $\min_z f(G(z))$, where the the generator $G$ is  nonlinear and $f$ is  convex. We then find the optimal \textit{latent vector} $z$, as illustrated in Section \ref{sec:experiments} with several examples. 
% such that $G(z)$ minimizes certain convex
%functional. 
%This gives rise to a poorly understood highly non-convex
%optimization problem due to the presence of the non-linear map $G$.

This GAN-based optimization problem poses various difficulties for existing first-order algorithms.  Indeed, to our knowledge, the only existing provable algorithm for solving \eqref{eq:emp loss min} relies  on  the existence of a projection oracle, and is limited to  the special case of \emph{compressive sensing} with a generative prior~\citep{Shah2018,Hegde2018}, see Section \ref{sec:related_works} for the details. 
%. More specifically, in the context of compressive sensing with a generative prior,  \citet{Shah2018,Hegde2018} proves linear convergence for the Projected Gradient Descent (PGD) in the presence of a projection oracle and in special case of $F(w,z)=\normb{Aw-b}_2^2$ for some \textit{measurement matrix}
%$A$ in \eqref{eq:emp loss min}. 
The main computational bottleneck is of course the non-convex projection step, %constraints set in \eqref{eq:emp loss min}, 
%oracle corresponding to the solution of \eqref{eq:emp loss min} with $F(w,z)
%=\normb{w - w^\natural}_2^2$. Even in this simple case, 
for which {no convergence analysis in terms of the geometry of the underlying  generator $G$ currently exists.}

On the other hand, Gradient Descent (GD) and its adaptive variants \citep{Kingma2014}  cannot efficiently handle non-smooth objective functions, as they are entirely oblivious to the composite structure of the problem \citep{nesterov2013gradient}. A simple example is denoising with the $\ell_\infty$-norm, for which subgradient descent (as the standard non-smooth alternative to GD) fails in practice, as observed in Section
\ref{sec:experiments}.
%In order
%to profit from the well-understood analysis of such c
%In analogy with composite convex optimization \citep{nesterov2013gradient},
%it is therefore imperative to design a flexible class of algorithms that efficiently handle non-smooth objectives.
%different breed of algorithms that {decouple} the
%convex objective function and the non-convex constraint in \eqref{eq:emp loss
%min}. 

With the explosion of GANs in popularity, there is consequently a pressing need
for provable and flexible optimization algorithms to solve the resulting
non-convex and (possibly) non-smooth problems. The present work addresses this need by focusing on the
general optimization template
\begin{align}
\begin{array}{lll}
    \underset{w, z}{\text{minimize}} & & F(w,z) := L(w) + R(w) + H(z) \\
\text{subject to } & & w = G(z),
\end{array}
\label{eq:emp loss min}
\end{align}
where $L:\R^d \rightarrow \R$ is  convex and smooth, $R:\R^d \rightarrow \R$ and
$H: \R^s \rightarrow \R$ are convex but not necessarily smooth, and
$G:\R^s\rightarrow\R^d$ is differentiable but often non-linear, corresponding to the generator network associated with  a GAN.
Even though $R$ and $H$ might not be smooth, we assume throughout that their
proximal mappings can be efficiently computed \citep{Parikh2014}.  

For brevity,
we refer to~\eqref{eq:emp loss min} as \textit{optimization with a generative
prior} whenever $G$ is given by the generator neural network associated with a GAN
\citep{Goodfellow2014}. In this context, we make three key contributions, summarized below:
%Our contributions can be summarized as

%is that they are oblivious to the
%particular composite structure of the objective, and might handle the possibly
%nonsmooth components $R$ and $H$ inefficiently. 

%Let us summarize the main contributions of this work. 

\textbf{1. Algorithm:} We propose an efficient and scalable (linearized)
Alternating Direction Method-of-Multipliers (ADMM) framework  to
solve \eqref{eq:emp loss min}, see Algorithm \ref{alg:admm}. To our
knowledge, this is the first non-convex and linearized ADMM algorithm for nonlinear constraints with provable fast
rates to solve problem \eqref{eq:emp loss min}, see
Section~\ref{sec:related_works} for a detailed literature review.

We evaluate this algorithm numerically in the context of denoising with GANs in
the presence of  adversarial or stochastic noise, as well as compressive
sensing~\citep{Bora2017}. In particular, Algorithm~\ref{alg:admm} allows for
efficient denoising with the $\ell_{\infty}$- and $\ell_1$-norms, with applications
in defenses against adversarial examples \citep{Szegedy2013} and signal
processing, respectively. 
%the numerical evidence suggests
%denoising with $\ell_\infty$ or $\ell_1$ norms is both efficient and also
%improves over iterative descent methods. 

\textbf{2. Optimization guarantees:} We prove  fast approximate convergence for
Algorithm~\ref{alg:admm} under the assumptions of smoothness and near-isometry
of $G$, as well as strong convexity of $L$.
%, as detailed in Section
%\ref{sec:guarantees1}.  % (supplementary \ref{sec:rsc}).
That is, we distill the key geometric attributes of the generative
network $G$  responsible for the success of Algorithm~\ref{alg:admm}.
We then show how some common neural network architectures 
satisfy these geometric assumptions. 

{We also establish a close relation between a variant of Algorithm~\ref{alg:admm} and the gradient descent in \citep{Bora2017} and, in this sense, provide the first rates for it, albeit in a limit case detailed in Section~\ref{sec:guarantees1}. Indeed, one key advantage of the primal-dual formulation studied in this paper is exactly this versatility, as well as the efficient handling of non-smooth objectives.} Lastly, we later relax the assumptions on $L$ to
\textit{restricted} strong convexity/smoothness, thus extending our results to
the broader context of statistical learning with generative priors, which includes
compressive sensing \citep{Bora2017} as a special case. 
%\label{pr:G-assumptions}.

\textbf{3. Statistical guarantees:} In the context of statistical learning with generative priors, where $L$ in~\eqref{eq:emp loss min} is replaced with an 
%$L(w)=\E_{x \sim \mu_m}
%l(w;x)$ 
\textit{empirical risk},
% where $\mu_m$ is an empirical
%distribution corresponding to a sample from $\mu$ and $\hat{L}(w)=\E_{x \sim
%\mu} l(w;x)$ is the \textit{population risk}, our main theoretical result is
%\ref{thm:main2} 
we provide the generalization error associated with Algorithm~\ref{alg:admm}. That is, we use the standard notion of Rademacher complexity~\citep{mohri2018foundations} to quantify the number of training data points required for Algorithm~\ref{alg:admm} to learn the true underlying parameter $w^\natural$.

\section{Algorithm}
\label{sec:algorithm}
In this section, we adapt the powerful Alternating Descent Method of
Multipliers (ADMM)~\citep{Glowinski1975,Gabay1976,Boyd2011} to solve the
non-convex {problem}~\eqref{eq:emp loss min}. We define the corresponding \emph{augmented Lagrangian} with the dual variable $\lambda
\in \R^p$ as
\begin{equation}
\begin{split}
    & \Lagr_\rho(w, z, \lambda) := L(w)+ \langle w - G(\prior), \lambda \rangle + \frac{\rho}{2}\norm{w - G(\prior)}_2^2,
\end{split}
\label{eq:defn_aug_lagr}
\end{equation}
for a penalty weight $\rho>0$.
By a standard duality argument, \eqref{eq:emp loss min} is equivalent to
\begin{align}
\underset{w,z}{\text{min}}\, \underset{\lambda}{\text{max}} \,
  \Lagr_\rho(w,z,\lambda)+ R(w)+H(z).
  \label{eq:aug lagr}
\end{align}
Applied to \eqref{eq:aug lagr}, every iteration of ADMM would minimize
the augmented Lagrangian with respect to $z$, then with respect to $w$, and then
update the dual variable $\lambda$. Note that $\Lagr_\rho(w, z, \lambda)$ is often
non-convex with respect to $z$ due to the {nonlinearity of the} generator $G:\R^s \rightarrow \R^d$
and, consequently, the minimization step with respect to $z$ in ADMM is often
intractable. 

To overcome this limitation, we next \emph{linearize} ADMM.
%\edita{This linearization finds precedent in \citet{liu2017linearized},
%but without any convergence rates. \marginnote{\textbf{Can we find any more
%references?}}} 
In the following, we let $\Proj_R$ and $\Proj_H$ denote the
\emph{proximal maps} of $R$ and $H$, respectively~\citep{Parikh2014}.

The equivalence of problems \eqref{eq:emp loss min} and \eqref{eq:aug
lagr} motivates us to consider the following algorithm for the penalty weight $\rho>0$, the primal
step sizes  $\alpha, \beta>0$, and the positive dual step sizes $\{\s_t\}_{t\ge
0}$:
\begin{equation}
\begin{split}
z_{ t+1} &= \Proj_{\b H} \l( z_{ t} - \b \nabla_{z} \Lagr_{\rho} (w_{t}, z_{t},  \lambda_{t}) \r), \\
    w_{ t+1} &= \Proj_{\a R} \left( w_{t} - \a \nabla_{w}
                        \Lagr_{\rho}(w_{t}, z_{t+1},  \lambda_{t}) \right), \\
    \lambda_{t+1} &= \lambda_{ t} + \sigma_{t+1}  (w_{t+1}-G(z_{t+1})).
\end{split}
\label{eq:aug lagr alg}
\end{equation}
As opposed to ADMM, to solve \eqref{eq:emp loss min}, the linearized ADMM in \eqref{eq:aug lagr alg} takes only
one descent step in both $z$ and $w$,
%The linearized ADMM developed above for
%solving \eqref{eq:emp loss min} 
see Algorithm \ref{alg:admm} for the summary. The particular choice of the dual step sizes $\{\s_t\}_t$ in
Algorithm~\ref{alg:admm} ensures that the dual variables $\{\lambda_t\}_t$
remain bounded,  see~\citep{bertsekas1976penalty} for a precedent in the convex
literature. 

{
\paragraph{Algorithm 2.} Let us introduce an important variant of Algorithm~\ref{alg:admm}. In our setting, $\Lagr_\rho(w,z,\lambda)$ is in fact convex with respect to
$w$ and therefore Algorithm 2 replaces the first step in \eqref{eq:aug lagr alg}
with exact minimization over $w$. {This} exact minimization step can be
executed with an off-the-shelf convex solver, or might sometimes have a closed-form solution. Moreover, Algorithm 2 gradually increases the penalty weight to emulate a multi-scale structure. More specifically, for an integer $K$, consider  the sequences of penalty weights and primal step sizes $\{\rho_k,\a_k,\b_k\}_{k=1}^K$, specified as 
\begin{align}
\rho_k =   2^k\rho,
\qquad 
\a_k = 2^{-k} \a, 
\qquad 
\b_k  =2^{-k} \b,
\qquad k\le K.
\end{align}
Consider also a sequence of integers $\{n_k\}_{k=1}^K$, where 
\begin{align}
n_k = 2^k n,
\qquad k\le K,
\end{align}
for an integer $n$. At (outer) iteration $k$, Algorithm~2 executes $n_k$ iterations of Algorithm~\ref{alg:admm} with exact minimization over $w$. Then it passes the current iterates of $w$, $z$, and dual step size to the next (outer) iteration. Loosely speaking, Algorithm~2 has a multi-scale structure, allowing it to take larger steps initially and then slowing down as it approaches the solution.  As discussed in Section~\ref{sec:guarantees1}, the theoretical guarantees for Algorithm~\ref{alg:admm} also apply to Algorithm~2. 
%Moreover, Algorithm~2 is closely related to gradient descent in \citep{Bora2017} and, in this sense, we establish the first rates for GD, albeit in the limit case discussed in Section~\ref{sec:guarantees1}.
}
%e also Supplementary~\ref{sec:relations} for the relation  with
%other  algorithms, and the applicability of our theoretical guarantees. %for them.

%In addition to Algorithm~\ref{alg:admm}, we consider a variant that gradually increases the penalty weight, which we will refer to as Algorithm 2. For an integer $K$, consider  the sequences of penalty weights and primal step sizes $\{\rho_k,\a_k,\b_k\}_{k=1}^K$, specified as 
%\begin{align}
%\rho_k =   2^k\rho,
%\qquad 
%\a_k = 2^{-k} \a, 
%\qquad 
%\b_k  =2^{-k} \b,
%\qquad k\le K.
%\end{align}
%Consider also a sequence of integers $\{n_k\}_{k=1}^K$, where 
%\begin{align}
%n_k = 2^k n,
%\qquad k\le K,
%\end{align}
%for an integer $n$. At (outer) iteration $k$, Algorithm~2 executes $n_k$ iterations of Algorithm~\ref{alg:admm} and then passes the current $w$, $z$, and dual step size to the next (outer) iteration. Loosely speaking, Algorithm~2 has a multi-scale structure, allowing it to take large steps initially and then refine its iterations as it approaches the solution. 
%

As the closing remark, akin to the convex case~\citep{he2000alternating,xu2017admm}, it is also possible to devise a variant of Algorithm~\ref{alg:admm} with adaptive primal step sizes, which we leave for a future work.

% In fact, the common fixed dual step size in convex optimization
% fails for our problem in practice. \edita{<---is the last sentence correct? }  

%To avoid tuning  $\rho$ we adaptively
%choose a penalty parameter for each iteration as follows: %, which we also use  in our convergence charcterizations:
%\begin{align*}
%\rho_t =
%\begin{cases}
%\rho_{t-1}, \qquad \text{if } \|w_t - G(z_t)\|_2 \le  \frac{\rho_{0}}{\sqrt{t\log^{{2}} (t+1)}},\\
%\\
%\rho_{t-1}
%\sqrt{\dfrac{(t+1)\log^{{2}}(t+2)}{t\log^{{2}} (t+1)}}, \qquad
%\text{otherwise}
%\end{cases}
%%\label{eq:beta rule}
%\end{align*}
%for $t \ge 1$ and initialization penalty $\rho_0 > 0$.

\begin{algorithm}
    \caption{Linearized ADMM for solving problem \eqref{eq:emp loss min}}
    \label{alg:admm}
    \textbf{Input: } Differentiable  $L$,
    proximal-friendly convex regularizers $R$ and $H$, differentiable  prior $G$,
    penalty weight $\rho>0$, primal step sizes $\alpha,\beta>0$, initial dual
    step size $\s_0>0$, primal initialization $w_0$ and $z_0$, dual
    initialization $\lambda_0$, stopping threshold $\tau_c>0$.
    \begin{algorithmic}[1]
    \For{$t=0,1,\ldots, T-1$}
    
        \State $z_{ t+1} \gets \Proj_{\b H} \l( z_{ t} - \b \nabla_{z} \Lagr_{\rho} (w_{t}, z_{t},  \lambda_{t}) \r)$ \Comment{primal updates}
   
        \State $w_{ t+1} \gets \Proj_{\a R} \left( w_{t} - \a \nabla_{w}
                        \Lagr_{\rho}(w_{t}, z_{t+1},  \lambda_{t}) \right)$
                        
        \State $\s_{t+1} \gets \min\left(\s_0, \dfrac{\s_0}{\|w_{t+1}-G(z_{t+1})\|_2 t\log^2(t+1)}   \right )$ \Comment{dual step size}
        \State $\lambda_{t+1} \gets \lambda_{t} + \sigma_{t+1}  (w_{t+1}-G(z_{t+1}))$ \Comment{dual update}

        \State $s \gets \dfrac{\norm{z_{t+1}-z_t}_2^2}{\a}+\dfrac{\norm{w_{t+1}-w_t}_2^2}{\b} + \s_t \norm{w_t - G(z_t)}_2^2 \le \tau_c$ \Comment{stopping criterion}
        \If {$s \leq  \tau_c$}
        \State
        \Return{$(w_{t+1}, z_{t+1})$}
        \EndIf
    \EndFor \\
    \Return{$(w_T,z_T)$}
    \end{algorithmic}
\end{algorithm}

\section{Optimization Guarantees}
\label{sec:guarantees1}
Let us study the theoretical guarantees of Algorithm \ref{alg:admm} for
solving program \eqref{eq:emp loss min}, whose constraints are nonlinear and non-convex (since $G$ is specified by a neural network). The main contribution of this section is Theorem \ref{thm:main}, which is
inherently an optimization result stating that Algorithm \ref{alg:admm}
succeeds under certain assumptions on 
\eqref{eq:emp loss min}. 

From an optimization perspective, to our knowledge, Theorem \ref{thm:main} is the first to provide (fast) rates for non-convex and linearized ADMM, see Section \ref{sec:related_works} for a detailed literature review. The  assumptions   imposed below on $L$ and
the generator $G$ ensure the success of Algorithm \ref{alg:admm} and are shortly justified for our setup, where $G$ is a generator network.
\begin{assumption}
\textbf{\emph{strong convexity / smoothness of $L$:}}
\label{assump:risk}
We assume that $L$ in (\ref{eq:emp loss min}) is both strongly convex and
smooth, namely, there exist $0<\mu_L\le \nu_L$ such that
%the population risk $L$ defined as
%\begin{equation}
%    \label{eq:poploss}
%L(w) := \E_{x \sim \chi} l(w,x),
%\end{equation}
%\end{comment}
\begin{align}
\frac{\mu_L}{2} \| w-w'\|^2&  \le L(w') - L(w) - \langle w'-w,\nabla L(w) \rangle  \le \frac{\nu_L}{2} \| w-w'\|^2,
\quad \forall w, w'\in \R^d.
\label{eq:L_str_cvx}
\end{align}
\end{assumption}
Assumption~\ref{assump:risk} is necessary to establish fast rates for Algorithm~\ref{alg:admm}, %, which is arguably the most important regime in practice. 
 and is readily met for
$L(w)=\|{  w-\widehat{w}}\|_2^2$ with $\mu_L = \nu_L = 1$, which renders Algorithm~\ref{alg:admm} applicable to $\ell_2$-denoising with generative prior  in~\citep{Tripathi2018,Samangouei2018,Ilyas2017}.  Here, $\widehat{w}$ is the noisy image.
%Loosely speaking, Assumption \ref{assump:risk} guides Algorithm~\ref{alg:admm} to reach the solution of problem~\eqref{eq:emp loss min}. 
%Recalling the setup in Section \ref{sec:introduction}, we observe that Assumption \ref{assump:risk} evidently holds for DARN with $\ell_2$, where $L(w)=\frac{1}{2}\| w^\natural - w\|_2^2$ and, consequently, $\mu_L = \nu_L = 1$.  

In Supplementary \ref{sec:statLearn}, we also relax the strong
convexity/smoothness in Assumption \ref{assump:risk} to \emph{restricted}
strong convexity/smoothness, which enables us to apply Theorem \ref{thm:main}
in the context of statistical learning with a generative prior, for example in
compressive sensing~\citep{Bora2017}.  Under Assumption~\ref{assump:risk}, even
though $L$ and consequently the objective function of \eqref{eq:emp loss min}
are strongly convex, problem \eqref{eq:emp loss min} might \emph{not} have a
unique solution, which is in stark contrast with convex optimization. Indeed, a
simple example is minimizing $x^2+y^2$ with the constraint $x^2+y^2=1$. We
next state our assumptions on the generator $G$.
%\begin{comment}
%Assumptions~\ref{assump:props_of_l} and \ref{assump:risk}
% are standard in statistical learning~\cite{mohri2018foundations}. For example, in linear regression, we might take
% \[
% l(w,x)= \frac{1}{2}|\langle w-w^\n,x \rangle|^2,
% \]
% \[
% L_m(w) = \frac{1}{2m} \sum_{i=1}^m |\langle w-w^\n,x_i \rangle|^2,
% \]
%for which both Assumptions \ref{assump:props_of_l} and \ref{assump:risk} are met.
%\end{comment}
\begin{assumption}
\label{assump:G_ssmth}
\textbf{\emph{Strong smoothness of $G$:}}
Let $DG$ be the Jacobian of $G$. We assume that $G:\R^s \to \R^d$ is strongly
    smooth, namely, there exists $\nu_G \ge 0$ such that 
\begin{align}
 \| G(z') - G(z) - DG(z)\cdot  (z'-z) \|_2 \le \frac{\nu_G}{2} \|z'-z\|_2^2,
 \qquad \forall z,z'\in \R^s,
\end{align}
\end{assumption}

\begin{assumption}
\label{defn:G_iso}
\textbf{\emph{Near-isometry of $G$:}}
We assume that the generative prior $G$ is a near-isometric map, namely, there exist
$0 <\iota_G\le \kappa_G$  such that 
\begin{align}
\iota_G \|z' - z\|_2  \le  \|G(z') - G(z)\|_2
\le  \kappa_G \|z' - z\|_2 ,
\qquad \forall z,z'\in \R^s.
\label{eq:lipsG}
\end{align}
\end{assumption}

The invertibility of  certain GAN architechtures have been established before
in \citep{ma2018invertibility,hand2017global}.  More concretely, Assumptions
\ref{assump:G_ssmth} and \ref{defn:G_iso} hold for a broad class of generators,
as summarized in Proposition \ref{pr:G-assumptions} and proved in Supplementary
\ref{sec:proposition}.

\begin{proposition}
\label{pr:G-assumptions}
Let $G_\Xi:\Dom \subset \R^d
\to \R^s$ be a feedforward neural network with weights $\Xi \in \R^h$, $k$
layers, non-decreasing layer sizes $s \leq s_1 \leq \ldots s_k \leq d$, with 
$\omega_i$ as activation function in the $i$-th layer, and compact domain
$\Dom$. For every layer $i$, suppose that the activation $\omega_i:\R \to \R$ is of class $C^1$ (continuously-differentiable)  and strictly increasing. 
%Let $G_\Xi:\Dom \subset \R^d
%\to \R^s$ be a feedforward neural network with parameters $\Xi \in \R^h$, $L$
%layers, non-decreasing layer sizes $s \leq s_1 \leq \ldots s_k \leq d$,
%$\omega_i$ as activation function in the $i$-th layer and compact domain
%$\Dom$. 
Then, after an arbitrarily small perturbation to the weights $\Xi$, Assumptions \ref{assump:G_ssmth} and \ref{defn:G_iso} hold 
almost surely  with respect to the Lebesgue measure.
\end{proposition}

A few comments about the preceding result are in order.

\textbf{Choice of the activation function: } Strictly-increasing $C^1$
activation functions in Proposition \ref{pr:G-assumptions}, such as the
Exponential Linear Unit (ELU) \citep{clevert2015fast} or
softplus \citep{Dugas2000}, achieve similar or better performance compared to the commonly-used (but
non-smooth) Rectified Linear Activation Unit (ReLU)~\citep{Xu2015,clevert2015fast,gulrajani2017improved,Kumar2017,kim2018memorization}.
%Even whet In the context of GANs,
%\citet{gulrajani2017improved} reports unsuccesful training with ELUs but
%succeeds at training with other $C^1$ and increasing  activation functions like softplus. 
%\edita{<--- not a good thing to say.}
%On the other hand, training GANs with ELUs has been succesful in
%\citep{Kumar2017} and \citep{kim2018memorization}. 
In our experiments in Section \ref{sec:experiments}, we found that using ELU
activations for the generator $G$ does not  
%noticeable 
adversely effect the representation power of the trained generator. Lastly,
the activation function for the final layer of the generator is typically
chosen as the sigmoid or tanh~\citep{Radford2015}, for which the conditions in
Proposition \ref{pr:G-assumptions} are also met.

\textbf{Compact domain: } The compactness
requirement in Proposition \ref{pr:G-assumptions} is mild.  Indeed, even though the
Gaussian distribution is the default choice as the input for the generator in a GAN,
training has also been successful using compactly-supported distributions,
such as the uniform distribution~\citep{Lipton2017}. Interestingly, even after
training with Gaussian noise, limiting the resulting generator to a
truncated Gaussian distribution can in fact boost the performance of GAN~\citep{brock2018large}, as measured with common metrics like the Inception
Score~\citep{Salimans2016} or
Frechet Inception Distance~\citep{Heusel2017}. This evidence suggests that
obtaining a good generator $G$ with compact domain is straightforward. In the
experiments of Section~\ref{sec:experiments}, we use truncated Gaussian on an Euclidean ball centered at the origin.
%the
%disk $\norm{z}_2 \leq r$ for some large $r > 0$.

\textbf{Non-decreasing layer sizes:} This is a standard feature of popular generator
architectures such as the DCGAN \citep{Radford2015} or infoGAN
\citep{Chen2016}. This property is also exploited in the analysis of the optimization
landscape of problem \eqref{eq:emp loss min} by \cite{hand2017global,Heckel2019}
and for showing invertiblity of (de)convolutional generators
\citep{ma2018invertibility}.

\paragraph{Necessity of assumptions on $G$:} Assumptions \ref{assump:G_ssmth}
and \ref{defn:G_iso} on the generator $G$ are necessary for the provable success of
Algorithm \ref{alg:admm}. Loosely speaking, Assumption \ref{assump:G_ssmth}
controls the curvature of the generative prior, without which the dual
iterations can oscillate without improving the objective. 

On the
other hand, the lower bound in \eqref{eq:lipsG} means that the generative prior
$G$ must be \emph{stably} injective: Faraway latent parameters should be mapped
to faraway outputs under $G$. As a pathological example, consider the
parametrization of a circle as $\{(\sin z,\cos z):z \in [0,2\pi)\}$.

This stable injectivity property in \eqref{eq:lipsG} is necessary for the
success of Algorithm \ref{alg:admm} and is not an artifact of our proof
techniques. Indeed, without this condition, the $z$ updates in Algorithm
\ref{alg:admm} might not reduce the feasibility gap $\norm{w-G(z)}_2$. Geometric assumptions on nonlinear
constraints have precedent in the optimization literature
\citep{birgin2016evaluation,flores2012complete,cartis2018optimality} and to a lesser extent in the literature of neural networks too
\citep{hand2017global,ma2018invertibility}, which we further discuss in Section
\ref{sec:related_works}.

Having stated and justified our assumptions on $L$ and the generator $G$ in
\eqref{eq:emp loss min}, we are now prepared to present the main technical
result of this section. Theorem \ref{thm:main} states that Algorithm
\ref{alg:admm} converges linearly to a small neighborhood of a solution, see
Supplementary \ref{sec:theory} for the proof. 

\begin{theorem}
\label{thm:main}
\textbf{\emph{(guarantees for Algorithm \ref{alg:admm})}}
Suppose that Assumptions~\ref{assump:risk}-\ref{defn:G_iso} hold. 
\begin{comment}
and $L_m$
satisfies the restricted strong convexity and smoothness in
Definition~\ref{defn:Lm_scvx} for a set $W\subset \R^p$ that contains a
solution $w^*$ of Program (\ref{eq:emp loss min}) and all the iterates
$\{w_t\}_{t\ge 0}$ of Algorithm \ref{alg:admm}.\footnote{If necessary, the inclusion
$\{w_t\}_{t\ge 0}\subset W$ might be enforced by adding the indicator function
of the convex hull of $W$ to $R$ in Program \eqref{eq:emp loss min}, similar
to \cite{agarwal2010fast}.}
\end{comment}
Let $(w^*,z^*)$ be a solution of program (\ref{eq:emp loss min}) and let $\lambda^*$ be a corresponding optimal dual variable.  Let also
$\{w_t,z_t,\lambda_t\}_{t\ge 0}$ denote the output sequence of
Algorithm~\ref{alg:admm}. 
Suppose that the primal step sizes $\a,\b$  satisfy
\begin{align}
\a  \le  \frac{1}{\nu_\rho},
%\widetilde{\alpha},%(\b,\rho,\s_l,\zeta_L,\s_L,\nu_G,\iota_G,\kappa_G,\mu_L,\ol{\mu}_L,\nu_L,\ol{\nu}_L,\lambda_0),
\qquad
\b \le \frac{1}{\xi_\rho + 2\a\tau_\rho^2}.
%\frac{\xi_\rho}{2}+\a \tau_\rho^2 \le \frac{1}{2\b}, 
%\widetilde{\beta},%(\rho,\s_l,\zeta_L,\s_L,\nu_G,\iota_G,\kappa_G,\mu_L,\ol{\mu}_L,\nu_L,\ol{\nu}_L,%\lambda_0),
%,(\a,\b,\rho,\s_l,\zeta_L,\s_L,\nu_G,\iota_G,\kappa_G,\mu_L,\ol{\mu}_L,\nu_L,\ol{\nu}_L,\lambda_0),
\qquad \s_0 \le \s_{0,\rho}.
\label{eq:step_size_reqs_thm}
\end{align}
Then  it holds that
\begin{align}
\frac{\|w_t-\wh{w}\|_2^2}{\a}+\frac{\|z_t-\wh{z}\|_2^2}{\b} \le 2(1-\eta_{\rho})^t \Delta_{0} +
\frac{\ol{\eta}_\rho}{\rho},
\label{eq:linr_cvg_thm}
\end{align}
\begin{align}
\label{eq:feas-thm}
\|w_t - G(z_t)\|_2^2 \le 
\frac{4(1-\eta_\rho)^t\Delta_0}{\rho}+ \frac{\widetilde{\eta}_\rho}{\rho^2}
,
\end{align}
for every iteration $t$. Above, $\Delta_0 = \mathcal{L}_\rho(w_0,z_0,\lambda_0)- \mathcal{L}_\rho(w^*,z^*,\lambda^*)$ is the initialization error, see~(\ref{eq:defn_aug_lagr}).
The convergence rate $ 1-\eta_\rho\in (0,1)$ and the quantities  $\nu_\rho,\xi_\rho,\tau_\rho,\s_{0,\rho},\ol{\eta}_\rho,\widetilde{\eta}_\rho$ above depend on the parameters in  Assumptions \ref{assump:risk}-\ref{defn:G_iso} and on $\lambda^*$,  as  specified in  the proof. As an example,  in the regime where $\mu_L \gg \rho$ and $\iota_G^2 \gg \nu_G$, we can take 
\begin{align*}
\a \approx \frac{1}{\nu_L}, \qquad \b\approx \frac{1}{\rho \kappa_G^2},
\qquad \frac{\rho \nu_G}{\kappa_G^2}\lesssim\s_0 \lesssim \rho\min \l(\frac{\mu_L^2}{\nu_L^2},\frac{\iota_G^4}{\kappa_G^4}\r),
\end{align*}
\begin{align}
\eta_\rho \approx \min\l(\frac{\mu_L}{\nu_L}, \frac{\iota_G^2}{\kappa_G^2}  \r),
\qquad 
\ol{\eta}_\rho \approx \widetilde{\eta}_\rho \approx  \max\l(\frac{\nu_L}{\mu_L}, \frac{\kappa_G^2}{\iota_G^2}  \r).
\label{eq:step-sizes-example}
\end{align}
Above, for the sake of clarity,   $\approx$ and $\lesssim$   suppress the universal constants, dependence on the initial dual  $\lambda_0$ and the corresponding step size $\s_0$. 
%Note that, in this regime, $\eta_\rho,\ol{\eta}_\rho,\widetilde{\eta}_\rho$ are nearly independent of $\rho$.
\end{theorem}
A few clarifying comments about Theorem \ref{thm:main} are in order. 

\textbf{Error:} According to Theorem \ref{thm:main}, if the primal and dual
step sizes are sufficiently small and Assumptions
\ref{assump:risk}-\ref{defn:G_iso} are met,  Algorithm \ref{alg:admm} converges
linearly to a \emph{neighborhood} of a solution $(w^*,z^*)$. The size of this
neighborhood depends on the penalty weight $\rho$ in \eqref{eq:defn_aug_lagr}.
For instance, in the example in Theorem~\ref{thm:main}, it is easy to verify that this neighborhood has a
radius of $O(1/\rho)$, which can be made smaller by increasing $\rho$. Theorem~\ref{thm:main} is however silent about
the behavior of Algorithm \ref{alg:admm} within this neighborhood. This is
to be expected. Indeed, even in the simpler convex case, where $G$ in program~\eqref{eq:emp loss min} would  have been an affine map, provably no first-order
algorithm could converge linearly to the solution
\citep{ouyang2018lower,agarwal2010fast}. 

Investigating the behavior of Algorithm \ref{alg:admm} within this
neighborhood, while interesting, arguably has  little practical value. For
example, in the convex case, ADMM would converge  slowly (sublinearly) in this neighborhood, which does not appeal to the practitioners.  As another example, when Algorithm~\ref{alg:admm} is applied in the context of statistical learning, there is no
benefit in solving~\eqref{eq:emp loss min} beyond the statistical accuracy of
the problem at hand \citep{agarwal2010fast}, see the discussion in Supplementary~\ref{sec:gen-err}. As such, we defer the study of the
local behavior of Algorithm \ref{alg:admm} to a future work. 

\textbf{Feasibility gap:} Likewise, according to \eqref{eq:feas-thm} in
Theorem \ref{thm:main}, the feasibility gap of Algorithm \ref{alg:admm} rapidly
reaches a plateau. In the example in Theorem \ref{thm:main}, the feasibility
gap rapidly reaches $O(1/\rho)$, where $\rho$ is the penalty weight in \eqref{eq:defn_aug_lagr}. As before, even in the convex case, no first-order algorithm could
achieve exact feasibility at linear rate~\citep{ouyang2018lower,agarwal2010fast}. 
%Understanding the sublinear local
%behavior of Algorithm \ref{alg:admm}, while interesting, is arguably of little
%practical interest and thus deferred to a future work. 
%\marginnote{\editf{it
%seems that the little practical interest argument is repeated here and in te
%preceding paragraph}} \marginnote{\textbf{\edita{what other aspects of theorem should we discuss? what did the reviewers pointed out?}}} 

\textbf{Intution:} While the exact expressions for the quantities in Theorem
\ref{thm:main} are given in Supplementary~\ref{sec:theory}, the example
provided in Theorem~\ref{thm:main} highlights the simple but instructive regime
where $\mu_L \gg \rho$ and $\iota_G^2 \gg \nu_G$, see Assumptions
\ref{assump:risk}-\ref{defn:G_iso}. 
Intuitively, $\mu_L \gg \rho$ means that minimizing the objective of~\eqref{eq:emp loss min} is prioritized over reducing the feasibility gap, see
\eqref{eq:defn_aug_lagr}. In addition,  $\iota_G^2 \gg \nu_G$ suggests that the
generative prior $G$ is very smooth. 

In this regime, the primal step size $\a$ for $w$ updates
is determined by how smooth $L$ is, and the primal step size $\b$ in the
latent variable $z$ is determined  by how smooth $G$ is, see
\eqref{eq:step-sizes-example}. Similar restrictions are standard in first-order
algorithms to avoid oscillations  \citep{nesterov2013introductory}. 
 
As discussed earlier, the algorithm rapidly reaches a neighborhood of size
$O(1/{\rho})$ of a solution and the feasibility gap plateaus at
$O(1/\rho)$. Note the  trade-off here for the choice of $\rho$: the larger the
penalty weight $\rho$ is, the more accurate Algorithm \ref{alg:admm} would be
and yet increasing $\rho$ is restricted by the assumption $\rho\ll \mu_L$.
Moreover, in this example, the rate $1-\eta_\rho$ of Algorithm~\ref{alg:admm}
depends only on the regularity of $L$ and $G$ in program \eqref{eq:emp loss
min}, see \eqref{eq:step-sizes-example}. Indeed, the more well-conditioned $L$
is and the more near-isometric $G$ is, the larger $\eta_\rho$ and the faster
the convergence would be.  

Generally speaking, increasing the penalty weight $\rho$ reduces the bias of Algorithm \ref{alg:admm} at the cost of a slower rate. 
Beyond our work, such dependence on the geometry of
the constraints has precedent in the literature of
optimization~\citep{birgin2016evaluation,flores2012complete,cartis2018optimality}
and manifold embedding theory~\citep{eftekhari2015new,eftekhari2017happens}. 

{\paragraph{Relation to simple gradient descent:} %, the variant of Algorithm \ref{alg:admm} introduced in Section \ref{sec:algorithm}, 
Consider a variant of Algorithm~\ref{alg:admm} that replaces the linearized
update for $w$ in \eqref{eq:aug lagr alg} with exact minimization with respect
to $w$, which can be achieved with an off-the-shelf convex solver or might have
a closed-form solution in some cases. The exact minimization over $w$ and
Lemma~\ref{lem:str cvx} together guarantee that Theorem~\ref{thm:main} also
applies to this variant of Algorithm~1. 

Moreover, as a special case of \eqref{eq:emp loss min} where $R\equiv 0$ and
$H\equiv 0$, this variant is closely related to GD~\citep{Bora2017}, presented
there without any rates. In Appendix~\ref{sec:gd-admm},  we establish that the
updates of both algorithms match as the feasibility gap vanishes. In this
sense, Theorem \ref{thm:main} provides the first rates for GD, albeit limited
to the limit case of vanishing feasibility gap. Indeed, one key advantage of
the primal-dual formulation studied in this paper is exactly this versatility
in providing a family of algorithms, such as Algorithms \ref{alg:admm} and 2,
that can be tuned for various scenarios and can also efficiently handle the
non-smooth case where $R$ or $H$ are nonzero in~\eqref{eq:emp loss min}. }

\section{Related Work}
\label{sec:related_works}
\cite{Bora2017}  empirically tune gradient descent for compressive sensing with a
{generative prior}  
\begin{align}
 \underset{z}{\min}\,\,
 \norm{A\cdot G(z)-b}^2_2, 
 %\qquad b:=Aw^\natural + \eta
\label{eq:intro_pr2}
\end{align}
which is a particular case of
template \eqref{eq:emp loss min} (without splitting). They also provide a
statistical generalization error dependent on a certain \textit{set restricted isometry property}
on the matrix $A$. More generally, Theorem \ref{thm:main_stat}
in Supplementary~\ref{sec:statLearn} provides statistical guarantees for
Algorithm \ref{alg:admm} using the standard  notion of empirical Rademacher
complexity~\citep{mohri2018foundations}.
%
%Despite the success of first order methods like
%gradient descent in minimizing such objective, much of the literature regarding
%this problem has focused on the statistical guarantees of either an exact or
%approximate solution of \eqref{eq:intro_pr2}. Theorem 1.1 and 1.2 in
%\citet{Bora2017} exploit the Lipschitzness of the generator $G$ and a certain
%\textit{set-restricted eigenvalue condition} to derive sample complexity
%estimates for the approximate recovery of the underlying true signal
%$w^\natural$. In stark contrast our derived statistical guarantee
%\ref{thm:main2} is given in terms of the standard statistical notion of
%empirical Rademacher complexity \citep{mohri2018foundations}, moreover such
%generalization bounds are given for the iterates of our algorithm
%\ref{alg:admm}.

\citet{hand2017global} analyze the optimization landscape of
\eqref{eq:intro_pr2} under the assumption that $G$ $(i)$ is composed of linear
layers and ReLU activation functions, $(ii)$ is sufficiently expansive at each
layer and $(iii)$ the network's weights have a Gaussian distribution or an
equivalent deterministic \textit{weight distribution condition}. Under such
conditions, they show global existence of descent directions outside small
neighborhoods around two points, but do not provide algorithmic convergence rates.  Their analysis requires ReLU activation in all layers of the
generator $G$, including the last one, which is  often not met in practice.
%whereas the common
%GANs trained in practice use sigmoid or Tanh.

On the other hand, our framework is not restricted to a particular network
architecture and instead isolates the necessary assumptions on the network $G$
for the success of Algorithm~\ref{alg:admm}. In doing so, we effectively
decouple the learning task from the network structure $G$ and study them
separately in Theorem \ref{thm:main} and Proposition \ref{pr:G-assumptions},
respectively. In particular, our theory in Section~\ref{sec:guarantees1} 
(Supplementary~\ref{sec:statLearn}) applies broadly to any nonlinear map $G$ that
meets Assumptions~\ref{assump:risk}-\ref{defn:G_iso} (Assumptions~\ref{assump:G_ssmth}-\ref{assump:risk_supp}), respectively. In turn, Proposition~\ref{pr:G-assumptions} establishes that the standard feed forward network with
common differentiable activation functions almost surely meets these assumptions. In this sense, let us also point to the work of~\citet{oymak2018sharp}, which is limited to linear regression with a nonlinear constraint, with its convex analogue studied
in~\citep{agarwal2010fast,giryes2016tradeoffs}. 

\citet{Heckel2019} provides a convergence proof for a modified version of
gradient descent, limited to \eqref{eq:intro_pr2} and without specifying a rate. We provide
the convergence rate for a  broad range of learning problems, and study the
statistical generalization. \citet{Hand2018} studied the \textit{phase
retrieval} problem, with a non-convex objective function that is not directly
covered by  \eqref{eq:emp loss min}. 

For the problem \eqref{eq:intro_pr2}, \citet{Shah2018, Hegde2018} proposed to
use Projected Gradient Descent (PGD) after splitting in a manner similar to our
template \eqref{eq:emp loss min}. If the projection (onto the range of the
prior $G$) is successful, and under certain additional conditions, the authors
establish linear convergence of PGD to a minimizer of \eqref{eq:intro_pr2}.
However, the projection onto the nonlinear range of $G$ is itself a
difficult non-convex program without any theoretical guarantees. In contrast,
we can solve the same problem without any projections while still providing a
convergence rate.

From an optimization perspective, there are no fast rates for linearized ADMM
with nonlinear constraints to our knowledge, but convergence to a first-order
stationary point and special cases in a few different settings have been
studied~\citep{liu2017linearized,shen2016square,chen2014convergence,qiao2016linearized}.
Let us again emphasize that
Assumptions~\ref{assump:G_ssmth}~and~\ref{defn:G_iso} extract the key
attributes of $G$ necessary for the success of Algorithm \ref{alg:admm}, which
is therefore not limited to a generator network.  It is also worth noting
another line of work that applies tools from statistical physics to inference
with deep neural networks, see~\citep{manoel2017multi,rezende2014stochastic}
and the references therein.

\section{Experiments}
\label{sec:experiments}
In this section we evaluate our algorithms for image recovery tasks with a GAN
prior. The datasets we consider are the CelebA dataset of face images \citep{Liu2015}
and the MNIST dataset of handwritten digits~\citep{Lecun2010}. We
train a generator $G$ with ELU activation functions~\cite{clevert2015fast}, in
order to satisfy Assumption~\ref{assump:G_ssmth}. The generators are trained
using the Wasserstein GAN framework \citep{Arjovsky2017}. For the CelebA
dataset we downsample the images to $64 \times 64$ pixels as
in~\citet{gulrajani2017improved} and we use the same residual architecture
\citep{He2015} for the generator with four residual blocks followed by a
convolutional layer. For MNIST, we use the same architecture as one
in~\cite{gulrajani2017improved}, which contains one fully connected layer
followed by three deconvolutional layers.

We recover images on the range of the generator $G$, by choosing $z^\star \in
\mathbb{R}^s$ and setting $w^\star:=G(z^\star)$ as the true image to be
recovered. This sets the global minimum of our objective functions at zero, and
allows us to illustrate and compare the convergence rates of various algorithms.

Our Algorithm \ref{alg:admm} mantains iterates $\{w_t, z_t\}_t$ where $w_t$
might not be feasible, namely, $w_t$ might not be in the range of $G$. As the goal
in the following tasks is to recover an element in the range of $G$ (feasible
points of \eqref{eq:emp loss min}), we plot the objective value at the point $G(z_t)$.

\textbf{Baseline. } We compare to the most widely-used algorithm in the current
literature, the gradient descent algorithm (GD) as used in \citep{Bora2017},
where a fixed number of iterations with constant step size are performed for
the function $L(G(z))$. We tune its learning rate to be as large
as possible without \textit{overshooting}. (See Supplementary \ref{sec:exp_setup}
for details on the hyperparameter tuning).

Our goal is to illustrate our theoretical results and highlight scenarios
where Algorithm \ref{alg:admm} can have better performance than GD in
optimization problems with a generative prior. Hence, we do not compare with
sparsity-prior based algorithms, such as LASSO \citep{Tibshirani1996}, or argue
about GAN vs.\ sparsity priors as in \cite{Bora2017}.

\textbf{Our algorithms. } We will use $(i)$ (linearized) 
ADMM (Algorithm \ref{alg:admm}), and ($ii$) ADMM with exact minimization
(Algorithm 2 a.k.a. EADMM), described in Section~\ref{sec:algorithm}. For both
ADMM and EADMM, we choose a starting iterate (random $z_0$ and $w_0=G(z_0)$) and initial
dual variable $\lambda_0=0$ (for GD we choose the same $z_0$ as initial
iterate).. We carefully track the objective function value vs.
computation time for a fair comparison.

\begin{figure}[t]
    \centering
    \begin{minipage}{0.24\textwidth}
        \includegraphics[scale=0.5]{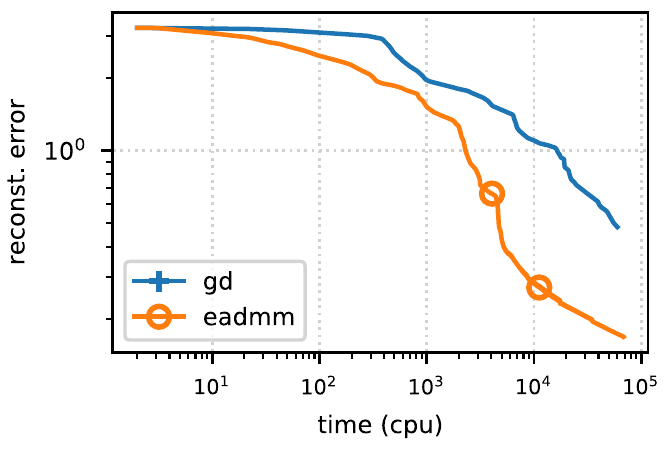}
    \label{fig:attack_error_time}
    \end{minipage} \hfill
    \begin{minipage}{0.24\textwidth}
        \includegraphics[scale=0.5]{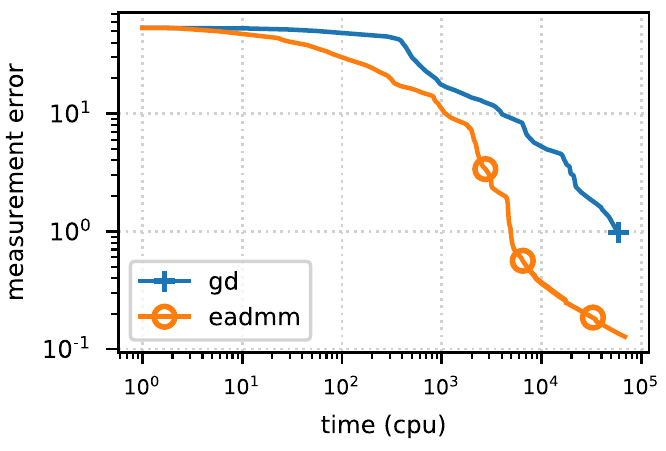}
    \label{fig:attack_error_time}
    \end{minipage} \hfill
    \begin{minipage}{0.24\textwidth}
        \includegraphics[scale=0.5]{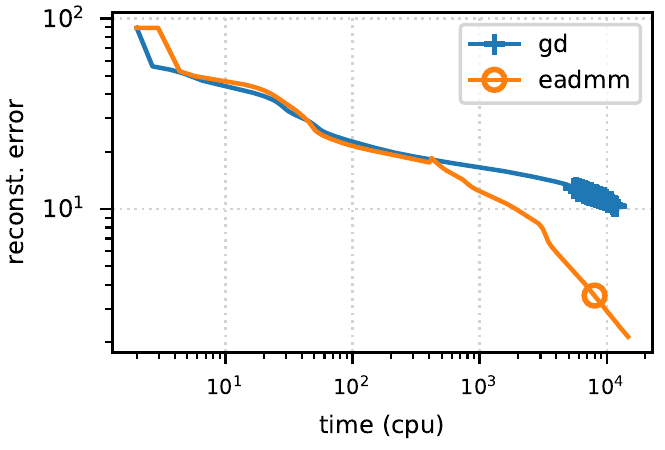}
    \label{fig:attack_error_time}
    \end{minipage} \hfill
    \begin{minipage}{0.24\textwidth}
    \centering
        \includegraphics[scale=0.5]{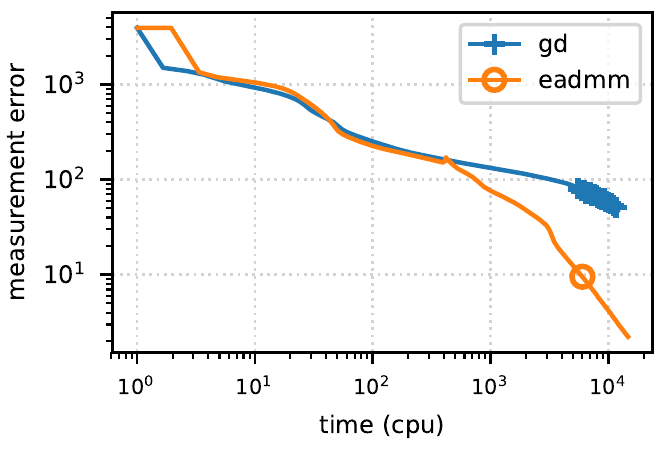}
    \label{fig:linf_comparison}
    \end{minipage}
    \caption{Reconstruction error and measurement error vs time. MNIST (left) and CelebA (right)}
    \label{fig:reconst}
\end{figure}

\textbf{Compressive sensing} The exact minimization step of EADMM  involves the
solution of a system of linear equations in each iteration. Performing Singular
Value Decomposition (SVD) once on the measurement matrix $A$, and storing its
components in memory, allows us to solve such linear systems with a very low
per-iteration complexity (see Supplementary~\ref{subsec:fast_exact}). We plot
the objective function value as well as the reconstruction error with $50\%$
relative measurements in Figure \ref{fig:reconst}(average over 20 images
(MNIST) and 10 images (CelebA)).

\textbf{Adversarial Denoising with $\ell_\infty$-norm}
\label{exp:ellinf}
Projection onto the range of a deep-net prior has been considered
by~\cite{Samangouei2018, Ilyas2017} as a defense mechanism against adversarial
examples~\citep{Szegedy2013}. In their settings, samples are denoised with a
generative prior, before being fed to a classifier. Even though the adversarial noise
introduced is typically bounded in $\ell_\infty$-norm, the projection is done
in $\ell_2$-norm. Such projection corresponds to $F(w,z)=\|w-w^\natural\|^2$
in~\eqref{eq:emp loss min}. We instead propose to project using the
$\ell_\infty$-norm that bounds the adversarial perturbation. To this end we
let $F(w, z) = \gamma\|w-w^\natural\|_2^2 + \|w-w^\natural\|_\infty$ in the
template~\eqref{eq:emp loss min}, for some small value of $\gamma$.
The proximal of the $\ell_\infty$ norm is efficiently
computable~\citep{Duchi2008}, allowing us to split $F(w, z)$ in its components
$L(w)=\gamma \|w-w^\natural\|_2^2$ and $R(w)=\|w-w^\natural\|_\infty$ (Note
that the small $\gamma$ ensures that Assumption~\ref{assump:risk} holds)
%\begin{wrapfigure}{r}{0.49\textwidth}
%    \centering
%        \includegraphics[scale=0.8]{\comparisonfolder/attack_comparison_error_time9.pdf}
%    \caption{Test error on denoised adversarial examples vs computation time.}
%    \label{fig:attack_error_time}
%\end{wrapfigure}
%
%\begin{wrapfigure}{r}{0.49\textwidth}
%    \centering
%        \includegraphics[scale=0.8]{\comparisonfolder/mnist_denoising_meas_-1.pdf}
%    \caption{$\ell_\infty$ reconstruction error per iteration for ADAM, GD, and EADMM.}
%    \label{fig:linf_comparison}
%\end{wrapfigure}

\begin{figure}
    \centering
    \begin{minipage}{0.49\textwidth}
        \includegraphics[scale=0.7]{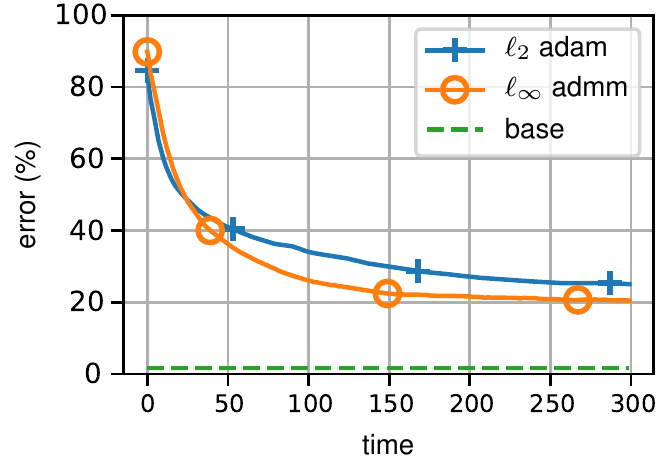}
    \caption{Test error on denoised adversarial examples vs computation time.}
    \label{fig:attack_error_time}
    \end{minipage} \hfill
    \begin{minipage}{0.49\textwidth}
    \centering
        \includegraphics[scale=0.7]{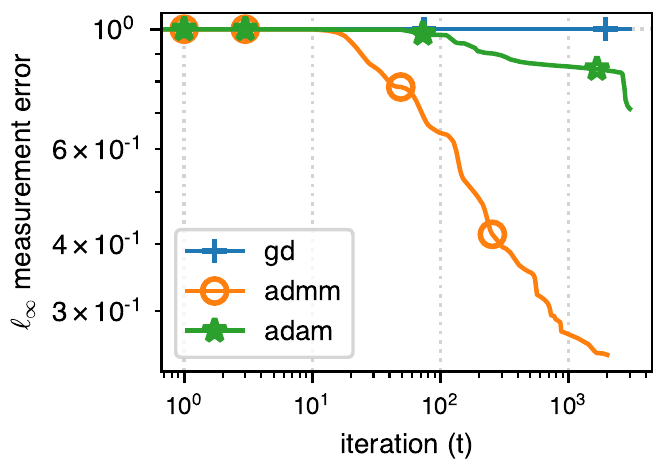}
    \caption{$\ell_\infty$ reconstruction error per iteration for ADAM, GD, and EADMM.}
    \label{fig:linf_comparison}
\end{minipage}
\end{figure}

We compare the ADAM optimizer \citep{Kingma2014},
GD and ADMM (450 iterations and for GD and ADAM, and 300 iterations for
EADMM). We use ADAM to solve the $\ell_2$ projection, while ADMM solves
the $\ell_\infty$ projection. We evaluate on a test set of 2000 adversarial
examples from the MNIST dataset, obtained with the Projected Gradient Algorithm
of \cite{Madry2018} with 30 iterations, stepsize 0.01 and attack size 0.2. For
the classifier, we use a standard convolutional network trained on clean MNIST
samples. We also test ADAM, GD (3000 iterations) and EADMM (2000 iterations)
on the $\ell_\infty$ denoising task.

The test error as a function of computation time is
in Figure~\ref{fig:attack_error_time}. We observe that the
$\ell_\infty$ denoising performs better when faced with $\ell_\infty$ bounded
attacks, in the sense that it achieves a lower error with less computation
time. In Figure~\ref{fig:linf_comparison}, we plot the $\ell_\infty$
reconstruction error achieved by ADAM, GD and EADMM, averaged over 7 images. GD
was unable to decrease the initial error, while ADAM takes a considerable
number of iterations to do so. In contrast, our ADMM already achieves the final
error of ADAM within its first 100 iterations.

\section*{Acknowledgements}
The authors have received funding for this project from the European Research
Council (ERC) under the European Union's Horizon 2020 research and innovation
programme (grant agreement no 725594 - time-data)
% \newpage

{\footnotesize
\bibliographystyle{plainnat}
\bibliography{gan_denoising}
}

\onecolumn
\appendix

%\section{Problem Formulation}
%\label{sec:problem}
%\input{sections/setup.tex}

%\section{Statistical Guarantees}
%\label{sec:appendix_guarantees2}
%\input{sections/guarantees2.tex}
%
%\section{Relation Between Algorithms}
%\label{sec:relations}
%\input{sections/relations.tex}

\section{Statistical Learning with Generative Priors}
\label{sec:statLearn}

So far, we have assumed $L$ to be strongly convex in \eqref{eq:emp loss min}, see Assumption \ref{assump:risk} and Theorem \ref{thm:main}. In this section, we relax this assumption on $L$ in the context of  statistical learning with generative priors, thus extending Theorem \ref{thm:main} to applications such as compressive sensing. We also provide the corresponding generalization error in this section. 

Here, we follow the standard setup in learning theory~\cite{mohri2018foundations}. 
Consider the probability space $(\mathbb{X}, \chi)$, where
$\mathbb{X} \subset \R^d$ is a compact set, equipped with the Borel sigma
algebra, and $\chi$ is the corresponding probability measure. To learn an
unknown parameter $w^\n \in \R^d$, consider the optimization program
\begin{align}
\underset{w\in \R^p}{\text{min}} \,\,  L(w),
\qquad
L(w) := \E_{x \sim \chi} l(w,x),
\label{eq:pop loss min}
\end{align}
where $L:\R^p\rightarrow \R$ is the differentiable \textit{population risk}
and $l:\R^d \times \R^p \rightarrow\R$ is the corresponding \emph{loss function}. We
also assume that Program \eqref{eq:pop loss min} has a unique solution $w^\n\in
\R^p$. The probability measure $\chi$ above is itself often unknown and we instead
have access to $m$ samples drawn independently from $\chi$, namely,
$\{x_i\}_{i=1}^m \sim \chi$. This allows us to form the \emph{empirical loss}
\begin{align}
L_m(w) := \frac{1}{m} \sum_{i=1}^m l(w,x_i).
\label{eq:emp loss defn supp}
\end{align}
Often, $m \ll p$ and to avoid an ill-posed problem, we must leverage any
inherent structure in $w^\n$. In this work, we consider a differentiable map
$G: \R^s \to \R^d$ and we assume that $w^\n \in G(\R^s)$.  That is,
there exists $z^\n \in \R^s$ such that $w^\n = G(z^\n)$. While not necessary, we limit ourselves in this section to the important case where $G$ corresponds to a  GAN, see Section \ref{sec:introduction}. 

%The defferentiability assumption on $G$ is not
%restrictive. For example, the non-differentiable ReLU activation function in a GAN can be  replaced with its smooth alternative ELU \citep{clevert2015fast}, which only differs
% from ReLU near the origin. In our experiments, we were able to succesfully train GANs with ELU activation function. \citet{gulrajani2017improved} also reports
%successful training of GANs with another differentiable activation function. See also the discussion after Assumption~\ref{assump:G_ssmth}. 

To learn $w^\n$ with the generative prior $w^\n=G(z^\n)$,  we propose
to solve the program
\begin{align}
\begin{array}{lll}
\underset{w,z}{\text{minimize}} & & L_m(w) + R(w) + H(z)\\
\text{subject to } & & w = G(z),
\end{array}
\label{eq:emp loss min supp}
\end{align}
where $R:\R^p \rightarrow \R$ and $H: \R^s \rightarrow \R$ are convex but not
necessarily smooth. Depending on the specific problem at hand, the \emph{regularizers}
$R$ and $H$ allow us to impose  additional structure on $w$ and
$z$, such as sparsity or set inclusion. Throughout, we again require that
the {proximal maps} \citep{Parikh2014} for $R$ and $H$ can be computed
efficiently, as detailed in Section \ref{sec:algorithm}. 

Let us now state our assumptions, some of which differ from Section \ref{sec:guarantees1}. 
\begin{assumption}
\label{assump:props_of_l_supp}
\textbf{\emph{Convexity / strong smoothness of loss: }}
We assume that $\lossf(\cdot, \cdot)$ is convex in both of its arguments.
Moreover, we assume that $\lossf(w, \cdot)$ is strongly smooth, namely, there
exists $\s_\lossf \ge 0$ such that for every $x,x'\in \mathbb{X}$
\begin{align}
    \D_\lossf(x,x';w) \le \dfrac{\s_\lossf}{2}\norm{x-x'}_2^2,
\label{eq:props_of_l_supp}
\end{align}
where $\D_\lossf$ stands for the \emph{Bregman divergence} associated with $l(w,\cdot)$,
\begin{align*}
\D_l(x,x';w) = l(w,x') - l(w,x) - \langle x'-x, \nabla_x l(w,x) \rangle .
\end{align*}
\end{assumption}
\begin{assumption}\textbf{\emph{Strong convexity / smoothness of the population risk:}}\label{assump:risk_supp}
We assume that the population risk $L$ defined as
\begin{equation}
    \label{eq:poploss_supp}
L(w) := \E_{x \sim \chi} l(w,x),
\end{equation}
is both strongly convex and smooth, i.e., there exist $0<\zeta_L\le \sigma_L$
such that
\begin{align*}
\frac{\zeta_L}{2} \| w-w'\|^2&  \le {D}_L(w,w') \le \frac{\sigma_L}{2} \| w-w'\|^2,
\end{align*}
\begin{align}
D_L(w,w') & = L(w') - L(w) - \langle w'-w,\nabla L(w) \rangle,
\label{eq:L_str_cvx_supp}
\end{align}
for every $w, w'\in \R^d$. In the following we denote by $w^\n$
the minimizer of (\ref{eq:poploss_supp}). In view of our assumption,
such minimizer is unique.
% NOTE BY FABIAN: here w^\n has not been properly introduced
% In particular, it follows that $w^\n$ is the unique
% minimizer of the population loss.
\end{assumption}
Assumptions~\ref{assump:props_of_l_supp} and \ref{assump:risk_supp}
 are standard in statistical learning~\cite{mohri2018foundations}. For example, in linear regression, we might take
 \[
 l(w,x)= \frac{1}{2}|\langle w-w^\n,x \rangle|^2,
 \]
 \[
 L_m(w) = \frac{1}{2m} \sum_{i=1}^m |\langle w-w^\n,x_i \rangle|^2,
 \]
for which both Assumptions \ref{assump:props_of_l_supp} and \ref{assump:risk_supp} are met. Lastly, we require that the Assumptions \ref{assump:G_ssmth} and \ref{defn:G_iso} on $G$ hold in this section, see and Proposition~\ref{pr:G-assumptions} for when these assumptions hold for generative priors. 

As a consequence of Assumption \ref{assump:props_of_l_supp}, we have that $L_m$ is
convex. We additionally require $L_m$ to be strongly convex and smooth in the
following restricted sense. Even though $L_m$ is random because of its
dependence on the random training data $\{x_i\}_{i=1}^m$, we ensure later in this section that the next condition is indeed met with high
probability when $m$ is large enough.
\begin{definition}
\label{defn:Lm_scvx}
\textbf{\emph{Restricted strong convexity / smoothness of empirical loss: }}
We say that $L_m$ is strongly convex and smooth on the set $W\subset\R^p$ if
there exist $0<\mu_L\le \nu_L$ and $\ol{\mu}_L,\ol{\nu}_L\ge 0$ such that
\begin{align}
\D_{L_m}(w,w') &   \ge \frac{\mu_L}{2}\|w'-w\|_2^2 - \ol{\mu}_L, \nonumber\\
\D_{L_m}(w,w')
& \le \frac{\nu_L}{2}\|w'-w\|_2^2 + \ol{\nu}_L,
\end{align}
\begin{align*}
\D_{L_m}(w,w') := L_m(w') - L_m(w) - \langle w' - w, \nabla L_m(w) \rangle,
\end{align*}
for every $w,w'\in W$.
\end{definition}
%Under the above assumptions, our first main result is that
%Algorithm \ref{alg:admm} converges linearly to a neighborhood of the solution, see Supplementary \ref{sec:theory} for the proof.

%While there are numerous quantities involved in the above assumptions and
%consequently in Theorem \ref{thm:main}, the main message is simple: The
%optimization error $\|w_t-w^*\|_2$ and the \emph{feasibility gap} $\|w_t - G(z_t)\|_2$
%converge linearly to reach a certain accuracy. Theorem \ref{thm:main} is
%immediately followed by a number of clarifying remarks.
%$\psi_\rho$ and $\mu_\rho \psi_\rho/\rho$ linearly, respectively.
%Here, $\rho$ is the penalty weight of the augmented Lagrangian and the quantities $\psi_\rho$ and $\mu_\rho$ are defined in the proof. 

Under the above assumptions, a result similar to Theorem \ref{thm:main} holds, which we state without proof.
\begin{theorem}
\label{thm:main-stat-learn}

\textbf{\emph{(guarantees for Algorithm \ref{alg:admm})}}
Suppose that Assumptions~\ref{assump:G_ssmth}-\ref{assump:risk_supp} hold. 
Let $(w^*,z^*)$ be a solution of program (\ref{eq:emp loss min}) and let $\lambda^*$ be a corresponding optimal dual variable.  Let also
$\{w_t,z_t,\lambda_t\}_{t\ge 0}$ denote the output sequence of
Algorithm~\ref{alg:admm}. 
Suppose that $L_m$
satisfies the restricted strong convexity and smoothness in
Definition~\ref{defn:Lm_scvx} for a set $W\subset \R^p$ that contains a
solution $w^*$ of  (\ref{eq:emp loss min}) and all the iterates
$\{w_t\}_{t\ge 0}$ of Algorithm \ref{alg:admm}.\footnote{If necessary, the inclusion
$\{w_t\}_{t\ge 0}\subset W$ might be enforced by adding the indicator function
of the convex hull of $W$ to $R$ in  \eqref{eq:emp loss min}, similar
to \cite{agarwal2010fast}.} Suppose also that the primal step sizes $\a,\b$ in  Algorithm~\ref{alg:admm} satisfy
\begin{align}
\a  \le  \frac{1}{\nu_\rho},
%\widetilde{\alpha},%(\b,\rho,\s_l,\zeta_L,\s_L,\nu_G,\iota_G,\kappa_G,\mu_L,\ol{\mu}_L,\nu_L,\ol{\nu}_L,\lambda_0),
\qquad
\b \le \frac{1}{\xi_\rho + 2\a\tau_\rho^2}.
%\frac{\xi_\rho}{2}+\a \tau_\rho^2 \le \frac{1}{2\b}, 
%\widetilde{\beta},%(\rho,\s_l,\zeta_L,\s_L,\nu_G,\iota_G,\kappa_G,\mu_L,\ol{\mu}_L,\nu_L,\ol{\nu}_L,%\lambda_0),
%,(\a,\b,\rho,\s_l,\zeta_L,\s_L,\nu_G,\iota_G,\kappa_G,\mu_L,\ol{\mu}_L,\nu_L,\ol{\nu}_L,\lambda_0),
\qquad \s_0 \le \s_{0,\rho}, 
\label{eq:step_size_reqs_thm}
\end{align}
Then  it holds that
\begin{align}
\frac{\|w_t-\wh{w}\|_2^2}{\a}+\frac{\|z_t-\wh{z}\|_2^2}{\b} \le 2(1-\eta_{\rho})^t \Delta_{0} +
\frac{\ol{\eta}_\rho}{\rho},
\label{eq:linr_cvg_thm}
\end{align}
\begin{align}
\label{eq:feas-thm}
\|w_t - G(z_t)\|_2^2 \le 
\frac{4(1-\eta_\rho)^t\Delta_0}{\rho}+ \frac{\widetilde{\eta}_\rho}{\rho^2}
,
\end{align}
for every iteration $t$. Above, $\Delta_0 = \mathcal{L}_\rho(w_0,z_0,\lambda_0)- \mathcal{L}_\rho(w^*,z^*,\lambda^*)$ is the initialization error, see~(\ref{eq:defn_aug_lagr}).
The convergence rate $ 1-\eta_\rho\in (0,1)$ and the quantities  $\nu_\rho,\xi_\rho,\tau_\rho,\s_{0,\rho},\ol{\eta}_\rho,\widetilde{\eta}_\rho$ above depend on the parameters in the Assumptions \ref{assump:G_ssmth}-\ref{assump:risk_supp} and on $\lambda_0,\s_0$. 
\end{theorem}

The remarks after Theorem \ref{thm:main} apply here too.

\subsection{Generalization Error \label{sec:gen-err}}
Building upon the optimization guarantee in Theorem~\ref{thm:main_stat}, our next result in this section is Theorem~\ref{thm:main_stat}, which quantifies the convergence of the iterates $\{w_t\}_{t\ge 0}$ of Algorithm~\ref{alg:admm} to the true parameter $w^\n$.

In other words, Theorem~\ref{thm:main_stat} below controls the generalization error of~\eqref{eq:emp loss min}, namely, the error  incurred by using the empirical risk $L_m$ in lieu of the population risk $L$. Indeed, Theorem~\ref{thm:main} is  silent about $\|w_t - w^\n\|_2$.  We address this shortcoming with the following result, proved in Section~\ref{sec:proof of ws_to_wn} of the supplementary material.  
\begin{lemma}\label{lem:ws_to_wn}
Let $R=1_W$ be the indicator function on $W\subset\R^p$ and set $H=0$ in ~(\ref{eq:emp loss min}).\footnote{To be complete, $1_W(w)=0$ if $w\in W$ and $1_W(w)=\infty$ otherwise. } Suppose that $w^*$ belongs to the relative interior of $W$. 
%\in \operatorname{relint}(W)$, where $\operatorname{relint}(W)$ denotes the relative interior of the set $W$.  
Then it holds that 
\begin{align}
\|w^\n - w^*\|_2 \le \frac{1}{\zeta_L} \max_{w\in W} \| \nabla L_m(w) - \nabla L(w) \|_2.
\label{eq:before_Upsilon}
\end{align}
\end{lemma}
Before bounding the right-hand side of \eqref{eq:before_Upsilon}, we remark that it is possible to extend Lemma~\ref{lem:ws_to_wn} to the case where the regularizer $R$ is a \emph{decomposable} norm, along the lines of~\cite{negahban2012unified}. We will however not pursue this direction in the present work. 
Next note that \eqref{eq:linr_cvg_thm} and Lemma~\ref{lem:ws_to_wn} together imply that 
\begin{align}
 \frac{\| w_t - w^\n\|_2^2}{\a^2} 
& \le \l( \frac{\|w_t - w^*\|_2}{\a}+ \frac{\|w^* - w^\n\|_2}{\b} \r )^2 
\qquad \text{(triangle inequality)} \nonumber\\
& \le \frac{2 \|w_t - w^*\|_2^2}{\a^2} + \frac{2 \|w^*-w^\n\|_2^2}{\b^2}
\quad ( (a+b)^2 \le 2a^2+2b^2  )
 \nonumber\\
& \le 
4(1-\eta_\rho)^t \Delta_0 + \frac{2\ol{\eta}_\rho}{\rho}
+ \frac{2}{\zeta_L^2} \max_{w\in W} \| \nabla L_m(w) - \nabla L(w)\|_2^2. 
\label{eq:before_rademacher}
\end{align}
According to Theorem~\ref{thm:main}, the right-hand side of \eqref{eq:before_rademacher} depends on $\mu_L,\ol{\mu}_L,\nu_L,\ol{\nu}_L$, which were introduced in Definition~\ref{defn:Lm_scvx}. 
%with the exact dependence specified in the proof of the theorem. 
Note that $\mu_L,\ol{\mu}_L,\nu_L,\ol{\nu}_L$  and the right-hand side of \eqref{eq:before_Upsilon} are all random variables because they depend on $L_m$ and thus on the randomly drawn training data $\{x_i\}_{i=1}^m$. To address this issue,  we apply a basic result in statistical learning theory as follows. For every $w\in \R^p$ and every pair $x,x'\in \mathbb{X}$, we use Assumption~\ref{assump:props_of_l_supp} to write that 
\begin{align}
\| \nabla l(w,x) - \nabla l(w,x') \|_2 & \le \s_l \| x-x'\|_2
\qquad \text{(see \eqref{eq:props_of_l_supp})} \nonumber\\
& \le \s_l \text{diam}(\mathbb{X}),
\end{align}
where $\text{diam}(\mathbb{X})$ denotes the diameter of the compact set $\mathbb{X}$. Note also that 
\begin{align}
\E_{\{x_i\}_i} \nabla L_m(w) = \nabla L(w), 
\qquad \forall w\in W,
\end{align}
where the expectation is over the training data $\{x_i\}_i$. 
Then, for $\varepsilon>0$ and except with a probability of at most $e^{-\varepsilon}$, it holds that 
\begin{align}
& \| \nabla L_m(w) - \nabla  L(w) \|_2  \nonumber\\
& \le 2 \mathcal{R}_W(x_1,\cdots,x_m) +3\s_l\text{diam}(\mathbb{X}) \sqrt{\frac{\varepsilon+2}{2m}} \nonumber\\
& =: \Upsilon_{m,W}(\varepsilon),
\label{eq:emp_proc}
\end{align}
for every $w\in W$~\citep{mohri2018foundations}. Above, 
\begin{align}
 \mathcal{R}_W(x_1,\cdots,x_m)  = 
\E_{E} \l[ \max_{w\in W}    \l \| \frac{1}{m} \sum_{i=1}^{m} e_i \nabla_w l(w,x_i)\r \|_2 \r],
\label{eq:emp rademach}
\end{align}
is the \emph{empirical Rademacher complexity} and $E=\{e_i\}_i$ is a Rademacher sequence, namely, a sequence of independent random variables taking $\pm 1$ with equal probabilities.  We can now revisit~\eqref{eq:before_rademacher} and write that 
\begin{align}
\|w_t- w^\n \|_2^2  & 
\le 4\a^2 (1-\eta_\rho)^t \Delta_0 + \frac{2\a^2 \ol{\eta}_\rho}{\rho}  + \frac{2\a^2\Upsilon_{m,W}^2(\varepsilon)}{\zeta_L^2},
\end{align}
which holds except with a probability of at most $e^{-\varepsilon}$. In addition, for every $w,w'\in W$, we may write that 
\begin{align}
& \| \nabla L_m(w) - \nabla L_m(w') \|_2  \nonumber\\
& \le   \| \nabla L(w) - \nabla L(w')\|_2+ \| \nabla L_m(w) - \nabla L(w) \|_2\nonumber\\
& \qquad  + \| \nabla L_m(w') - \nabla L(w')\|_2
\quad \text{(triangle inequality)} \nonumber\\
& \le  \s_L\|w-w'\|_2+  2\Upsilon_{m,W}(\varepsilon),
\qquad \text{(see (\ref{eq:L_str_cvx_supp},\ref{eq:emp_proc}))}
\end{align} 
except with a probability of at most $e^{-\varepsilon}$. 
Likewise, for every $w,w'\in W$, we have that 
\begin{align}
& \| \nabla L_m(w) - \nabla L_m(w') \|_2  \nonumber\\
& \ge \| \nabla L_m(w) - \nabla L_m(w) \|_2  -  \| \nabla L_m(w) - \nabla L(w)\|_2\nonumber\\
& \qquad - \| \nabla L_m(w') - \nabla L(w')\|_2
\qquad \text{(triangle inequality)} \nonumber\\
& \ge  \zeta_L\|w-w'\|_2-  2\Upsilon_{m,W}(\varepsilon),
\qquad \text{(see (\ref{eq:L_str_cvx_supp},\ref{eq:emp_proc} ))}
\end{align} 
except with a probability of at most $e^{-\varepsilon}$. Therefore, $L_m$ satisfies the restricted strong convexity and smoothness in Definition~\ref{defn:Lm_scvx} with
\begin{align}
\mu_L = \s_L,
\qquad 
\nu_L = \zeta_L,
\nonumber\\
\ol{\mu}_L = \ol{\zeta}_L =  2\Upsilon_{m,W}(\varepsilon). 
\end{align}
%Suppose that Assumptions~\ref{defn:Lm_scvx}-\ref{defn:G_iso} hold. These findings are summarized below in our second main result.  For a convex set $W\subset\R^p$, let  $R=1_W$  be the (convex) indicator function on $W$ and set $H=0$ in ~(\ref{eq:emp loss min}). Suppose that solution $w^*$ of ~(\ref{eq:emp loss min}) satisfies $w^*\in \operatorname{relint}(W)$, where $\operatorname{relint}(W)$ is the relative interior of the set $W$.  
Our findings in this section are summarized below. 
\begin{theorem} \label{thm:main_stat}\textbf{\emph{(generalization error)}} 
Suppose that Assumptions~\ref{assump:G_ssmth}-\ref{assump:risk_supp} hold and recall that the training samples $\{x_i\}_{i=1}^m$ are drawn independently from the probability space $(\mathbb{X},\chi)$ for a compact set $\mathbb{X} \subset \R^d$  with diameter $\operatorname{diam}(\mathbb{X})$.   

For a set $W\subset\R^p$, let  $R=1_W$  be the indicator function on $W$, and set $H\equiv 0$ in ~(\ref{eq:emp loss min}). Suppose that solution $w^*$ of ~(\ref{eq:emp loss min}) belongs to the relative interior of $W$. 
For $\varepsilon>0$, evaluate the quantities in Theorem~\ref{thm:main-stat-learn} with 
\begin{align}
\mu_L & = \s_L,
\qquad 
\nu_L = \zeta_L, \nonumber\\
\ol{\mu}_L & = \ol{\zeta}_L =   
4 \mathcal{R}_W(x_1,\cdots,x_m) \nonumber\\
& \qquad +6\s_l\operatorname{diam}(\mathbb{X}) \sqrt{\frac{\varepsilon+2}{2m}},
\end{align}
where $\mathcal{R}_W(x_1,\cdots,x_m)$ in the empirical  Rademacher complexity defined in (\ref{eq:emp rademach}). If the requirements on the step sizes in~(\ref{eq:step_size_reqs_thm}) hold, we then have that 
\begin{align}
\|w_t- w^\n \|_2^2  & 
\le   4\a^2(1-\eta_\rho)^t \Delta_0 + \frac{2\a^2 \ol{\eta}_\rho}{\rho}+ \frac{8\a^2 }{\zeta_L^2} \mathcal{R}_W(x_1,\cdots,x_m)^2  \nonumber\\
& \qquad +  \frac{18\a^2\s_l^2 \operatorname{diam}(\mathbb{X})^2( \varepsilon+2)}{m}, 
\end{align}
except with a probability of at most $e^{-\varepsilon}$. 
\end{theorem}
Most of the remarks about Theorem~\ref{thm:main} also apply to Theorem~\ref{thm:main_stat} and we note that $\|w_t-w^\n\|_2$ reduces by increasing the number of training samples $m$, before asymptotically reaching the generalization error
\begin{align}
2\psi_\rho + \frac{8 }{\zeta_L^2} \mathcal{R}_W(x_1,\cdots,x_m)^2.
\label{eq:statistical ball}
\end{align}
Computing the Rademacher complexity above for specific choices of the network structure and loss is itself potentially a complicated task, which we will not pursue by the virtue of the generality of our results so far. The key technical challenge there is computing the corresponding \emph{entropy integral}, which involves estimating the \emph{covering numbers} of the set $W$ \cite{mohri2018foundations}. One last takeaway point from the statistical accuracy in \eqref{eq:statistical ball} is the following. If  
\begin{align}
\ol{\eta}_\rho =O(\rho \cdot \mathcal{R}_W(x_1,\cdots,x_m)^2/\zeta_L^2),
\label{eq:statistacll ball 2}
\end{align}
the asymptotic optimization error  in Theorem~\ref{thm:main} does not play an important role in determining the generalization error above. In words, if  \eqref{eq:statistacll ball 2} holds, then Algorithm~\ref{alg:admm} converges to the ball of statistical accuracy around $w^\n$. Here, $O$ stands for the standard Big-O notation.

\section{Proof of Proposition \ref{pr:G-assumptions}}
\label{sec:proposition}
The feedforward network $G=G_\Xi:\R^s
\to \R^d$ is a composition of linear maps and entry-wise applications of
the activation functions, and hence is also of class $C^1$. Its Jacobian $DG:\R^s \to \R^{d \times s}$
is thus a continuous function and its restriction
to the compact subset $\Dom \subseteq \R^s$ is Lipschitz-continuous. Therefore, there exists
$\nu_G \geq 0$ such that
\[
    \norm{DG(z') - DG(z)}_2 \leq \nu_G \norm{z'-z}, \qquad \forall z, z' \in \Dom.
\]
From standard arguments it then follows that Assumption \ref{assump:G_ssmth} holds in the sense that 
\begin{align*}
\norm{G(z') - G(z) - DG(z)(z'-z)}_2 &= \norm{\int_{0}^1 (DG(tz' + (1-t)z) - DG(z))(z'-z) dt}_2 \\
&\leq \int_{0}^1 \norm{DG(tz' + (1-t)z) - DG(z)}_2 \norm{z' - z}_2 dt \\
&\leq \nu_G \int_0^1 t\norm{z' - z}^2 dt = \dfrac{\nu_G}{2} \norm{z'-z}_2^2,
\end{align*}
for every $z,z'\in \R^s$. 

In order to show that Assumption \ref{defn:G_iso} (near-isometry) also holds, we will require the following simple fact:

\begin{lemma}
\label{lem:inv_lips}
Let $G:\Dom \subseteq \R^s \to \R^d$ have a left inverse $H:G(\Dom) \subseteq \R^d \to \R^s$
which is Lipschitz-continuous with constant $\iota_G > 0$. Then  it holds
that
\[
    \dfrac{1}{\iota_G} \norm{z'-z} \leq \norm{G(z') - G(z)},
    \qquad \forall z', z \in \D.
\]
\begin{proof}
\[
\norm{z' - z} = \norm{H(G(z')) - H(G(z))} \leq \iota_G \norm{G(z') - G(z)}. 
\]
\end{proof}
\end{lemma}
We now proceed to show that Assumption \ref{defn:G_iso} holds. We suppose $G$ is of the form
\[
    G(z) = \omega_k W_k (\omega_{k-1} W_{k-1}\ldots (\omega_1 W_1 z)\ldots),
\]
for weight matrices $\{W_k\}_k$. 
First note that, by the compactness of the domain of $G$, 
the values of the hidden layers are always contained in a product of compact
intervals, and so we can replace $\omega_i$ by its restriction to such sets.
Each $\omega_i$ is continuous, defined on a product of intervals, and
is stricly increasing so that they have a continuous left inverse $\omega_i^{-1}$
\citep[Proposition 6.4.5]{garling_2014}. The
assumption of non-decreasing layer sizes implies that the $W_i$ are tall
matrices of dimensions $(m_i, n_i)$ with $m_i \geq n_i$, whose columns are
almost surely linearly independent after an arbitrarily small perturbation. In such case they have a
left matrix inverse $W_i^{-1}$, which as a {bounded} linear map, is continuous.
It then follows that $G$ has a continuous left inverse of the form
\[
    G^{-1}=W_1^{-1} \circ \omega_1^{-1} \ldots W_k^{-1} \circ \omega_k^{-1},
\]
which is a continuous mapping and is defined on $G(\Dom)$ which by continuity
of $G$ is compact, hence $G^{-1}$ is Lipschitz-continuous. The result then
follows by the Lipschitz continuity of the map $G$ (restricted to the compact domain $\Dom$)
and Lemma \ref{lem:inv_lips}.

\section{Proof of Theorem \ref{thm:main}}
\label{sec:theory}
It is convenient throughout the supplementary material  to use a slightly different notation for Lagrangian, compared to the body of the paper. 
To improve the readability of the proof, let us  list here the assumptions on the empirical loss $L$ and prior $G$ that are used throughout this proof. For every iteration $t$, we assume that 
\begin{align}
& L(w_{t}) - L(\wh{w}) - \langle w_{t}-\wh{w} , \nabla L(\wh{w}) \rangle \nonumber\\
& \ge \frac{\mu_L}{2}\|w_{t}-\wh{w}\|_2^2 , 
\qquad \l(\text{strong convexity of }L  \r)
\label{eq:str cvx of L}
\end{align}
\begin{align}
& L(w_{t+1}) - L(w_{t}) - \langle w_{t+1}-w_{t} , \nabla L(w_{t}) \rangle  \nonumber\\
& \le \frac{\nu_L}{2}\|w_{t+1}-w_{t}\|_2^2 , 
\qquad \l(\text{strong smoothness of } L \r)
\label{eq:smoothness of L}
\end{align}
\begin{align}
& \| G(z') - G(z) - DG(z) \cdot (z'-z) \|_2 \nonumber\\
& \le \frac{\nu_G}{2} \|z'-z\|_2^2,
\qquad \l(\text{strong smoothness of } G\r)
\label{eq:smoothness of G} 
\end{align}
\begin{align}
{\iota}_G \|z' - z\|_2   \le  \|G(z') - G(z)\|_2 \le \kappa_G \|z'- z\|_2,
\qquad \l( \text{near-isometry of } G \r)
\label{eq:lipschitz of G 2}
\end{align}
%\begin{align}
%\|G(z')-G(z)\|_2 \le \kappa_G \|z'-z\|_2+\ol{\kappa}_G,
%\qquad \l( \text{Lipschitz continuty of } G \r)
%\label{eq:lipschitz of G}
%\end{align}
\begin{align}
\|DG(z)\cdot (z'-z)\|_2 \le \kappa_G \|z'-z\|_2,
\qquad \l( \text{Lipschitz continuty of } G \r)
\label{eq:smoothness of G 2}
\end{align}
For the sake of brevity, let us set 
\begin{align*}
v = (w,z) \in \R^{p+s},
\end{align*}
\begin{align}
& \Lagr_\rho(v,\lambda) := \Lagr_\rho(w,z,\lambda)   := L(w) + R(w)+H(z)+ \langle w - G(z), \lambda \rangle \nonumber\\
& \qquad \qquad \qquad \qquad \qquad + \frac{\rho}{2}\|w-G(z)\|_2^2, 
\qquad \text{(augmented Lagrangian)}
\label{eq:ind Lagr}
\end{align}
\begin{align}
\Lagr_\rho'(v,\lambda) := \Lagr'_\rho(w,z,\lambda) = L(w)+\langle w - G(z),\lambda\rangle + \frac{\rho}{2}\|w-G(z)\|_2^2, 
\label{eq:ind Lagr prime}
\end{align}
\begin{align}
A(v) = A(w,z) := w - G(z).
\qquad \text{(feasibility gap)}
\label{eq:A_shorhtand}
\end{align}
Let also $v^*=(\wh{w},\wh{z})$ be a solution of \eqref{eq:emp loss min} and let $\wh{\lambda}$ be a corresponding optimal dual variable. The first-order necessary optimality conditions for \eqref{eq:emp loss min} are 
\begin{align}
\begin{cases}
- \nabla_v \Lagr'_\rho(\wh{v},\wh{\lambda}) \in \partial R(w^*)\times \partial H(z^*),\\
\wh{w} = G(\wh{z}),
\end{cases}
\label{eq:opt cnds}
\end{align}
where $\partial R(w^*)$ and $\partial H(z^*)$ are the  subdifferentials of $R$ and $H$, respectively, at $w^*$ and $z^*$.  
Throughout the proof, we will also often use the notation
\begin{align}
\Delta_{t} := \Lagr_\rho(v_{t},\lambda_{t}) - \Lagr_\rho(\wh{v},\wh{\lambda}), 
\label{eq:defn of Delta}
\end{align}
\begin{align}
\Delta'_t := \Lagr'_\rho(v_t,\lambda_t) - \Lagr'_\rho(v^*,\lambda^*),
\end{align}
\begin{align}
\delta_{t} := \| w_{t}-w^*\|_2,
\qquad 
\delta'_t := \| z_t - z^*\|_2,
\label{eq:defn of delta}
\end{align}
\begin{align}
A_{t} := A(v_t) = w_{t} - G(z_{t}).
\label{eq:defn of A}
\end{align}
In particular, with this new notation, the dual update can be rewritten as 
\begin{align}
\lambda_{t+1} = \lambda_t + \sigma_{t+1} A_{t+1}.
\qquad \text{(see Algorithm \ref{alg:admm})}
\label{eq:dual update proof}
\end{align}
First, in Appendix \ref{sec:smoothness}, we control the smoothness of $\Lagr'_\rho$ over the trajectory of the algorithm.
\begin{lemma}\label{lem:smoothness}
For every iteration $t$, it holds that 
\begin{align}
&  \Lagr'_\rho (w_{t+1},z_{t+1},\lambda_{t}) - \Lagr'_\rho (w_{t},z_{t+1},\lambda_{t}) - 
\langle w_{t+1}-w_{t}, \nabla_w \Lagr'_\rho(w_{t},z_{t+1},\lambda_{t})  \rangle \nonumber\\
& \le \frac{\nu_\rho  }{2}\|w_{t+1}-w_t\|_2^2,
\label{eq:smoothness lemma w}
\end{align}
\begin{align}
& \Lagr'_\rho (w_{t},z_{t+1}\lambda_{t}) - \Lagr'_\rho (w_{t},z_{t},\lambda_{t}) - 
\langle z_{t+1}-z_{t}, \nabla_z \Lagr'_\rho(w_{t},z_{t},\lambda_{t})  \rangle \nonumber\\
& \le \frac{\xi_\rho }{2}\|z_{t+1}-z_t\|_2^2,
\label{eq:smoothness lemma z}
\end{align}
\begin{align}
\| \nabla_w \Lagr'_\rho(w_t,z_{t+1},\lambda_t) - \nabla_w \Lagr'_\rho(w_t,z_{t},\lambda_t)  \|_2 & \le \tau_\rho \|z_{t+1}-z_t\|_2^2 ,
\label{eq:smoothness lemma zw}
\end{align}
where 
\begin{align}
\nu_\rho := \nu_L+\rho.
\end{align}
\begin{align}
\xi_\rho  := \nu_G (\lambda_{\max}+\rho \max_i \|A_i\|_2 ) + 2\rho \kappa_G^2,   
\label{eq:xi rho bar raw}
\end{align}
\begin{align}
\tau_\rho := \rho \kappa_G.
\end{align}
\end{lemma}
Second, in the following result we ensure that $\Lagr_\rho$ and $\Lagr_\rho'$ are sufficiently regular along the trajectory of our algorithm, see  Appendix  \ref{sec:str cvx} for the proof. 
%, as a consequence of (\ref{eq:smoothness of L},\ref{eq:smoothness of G},\ref{eq:lipschitz of G 2}).
\begin{lemma}\label{lem:str cvx}
For every iteration $t$, it holds that 
\begin{align}
\Delta_{t} \ge \frac{\mu_\rho \d_t^2}{2} + \frac{\mu_\rho' \d_t'^2}{2}  - \ol{\mu}_\rho,
\label{eq:str cvx lemma 1}
\end{align}
\begin{align}
 \Delta'_{t}+ \langle v^* - v_{t} , \nabla_v \Lagr'_\rho (v_{t}) \rangle 
\le  \frac{\omega_\rho \d_t^2}{2} + \frac{\omega'_\rho \d_t^{'2}}{2},
\label{eq:smoothness lemma 2}
\end{align}
where 
\begin{align}
\mu_\rho :=  \mu_L - 2\rho,
\qquad 
\mu'_\rho := \frac{\rho \iota^2_G}{2}-\nu_G\|\lambda^*\|_2,
\end{align}
\begin{align}
\ol{\mu}_\rho := 
\frac{3}{\rho}\l( \lambda_{\max}^2 + \|\lambda^*\|_2^2\r), 
\end{align}
\begin{align}
\omega_\rho := 0,
\qquad 
\omega'_\rho :=
\frac{\nu_G  }{2} \l(\lambda_{\max}+ \rho   \r).
\end{align}
\end{lemma}
Having listed all the necessary technical lemmas above, we now proceed to prove Theorem \ref{thm:main}. Using the smoothness of $\Lagr'_\rho$, established in Lemma \ref{lem:smoothness}, we argue that 
\begin{align}
& \Lagr'_\rho(v_{t+1},\lambda_{t+1}) \nonumber\\
& = L(w_{t+1})+ \langle A_{t+1}, \lambda_{t+1} \rangle + \frac{\rho}{2}\| A_{t+1}\|_2^2 
\qquad \text{(see \eqref{eq:ind Lagr prime})}
\nonumber\\
& = L(w_{t+1})+ \langle A_{t+1}, \lambda_{t} \rangle + \l(\frac{\rho}{2}+ \sigma_{t+1} \r) \| A_{t+1}\|_2^2 
\qquad \text{(see \eqref{eq:dual update proof})}
\nonumber\\
& = \Lagr'_\rho(w_{t+1},z_{t+1},\lambda_{t}) +  \sigma_{t+1}  \| A_{t+1}\|_2^2 
\qquad \text{(see (\ref{eq:ind Lagr}))}
\nonumber\\
& \le \Lagr'_\rho (w_t,z_{t+1},\lambda_t) + \langle w_{t+1}-w_t, \nabla_w \Lagr'_\rho(w_t,z_{t+1},\lambda_t) \rangle +  \frac{\nu_\rho}{2}\|w_{t+1}-w_t\|_2^2 \nonumber\\
& \qquad + \ol{\nu}_\rho + \s_{t+1}\|A_{t+1}\|_2^2 
\qquad \text{(see (\ref{eq:smoothness lemma w}))}\nonumber\\
& \le \Lagr'_\rho (w_t,z_{t+1},\lambda_t) + \langle w_{t+1}-w_t, \nabla_w \Lagr'_\rho(w_t,z_{t+1},\lambda_t) \rangle +  \frac{1}{2\a}\|w_{t+1}-w_t\|_2^2\nonumber\\
& \qquad  + \ol{\nu}_\rho+ \s_{t+1}\|A_{t+1}\|_2^2  ,
\label{eq:proof 0}
\end{align}
where the last line above holds if the step size $\a$ satisfies 
\begin{align}
\a \le \frac{1}{\nu_\rho}.
\label{eq:first-step-size-assump}
\end{align}
According to Algorithm~\ref{alg:admm}, we can equivalently write the $w$ updates as 
\begin{align}
w_{t+1} = \arg\min_w \l\langle w - w_{t}, \nabla_w \Lagr'_\rho(w_{t},z_{t+1},\lambda_{t})  \r\rangle + \frac{1}{2\a}\|w-w_{t} \|_2^2+ R(w).
\label{eq:equiv rep of w update}
\end{align}
In particular, consider above the choice of $w=\t w^*+(1-\t) w_t$ for $\t\in [0,1]$ to be set later. We can then bound the last line of \eqref{eq:proof 0} as 
\begin{align}
& \Lagr'_\rho(v_{t+1},\lambda_{t+1}) + R(w_{t+1}) \nonumber\\
& = \Lagr'_\rho(w_{t},z_{t+1},\lambda_{t})  +  
\min_{w}  \langle w - w_{t}, \nabla_w \Lagr'_\rho(w_{t},z_{t+1},\lambda_{t}) \rangle \nonumber\\
& \qquad +  \frac{1}{2\a}\|w-w_{t} \|_2^2 + R(w) + \sigma_{t+1} \|A_{t+1}\|_2^2
\qquad \text{(see (\ref{eq:proof 0},\ref{eq:equiv rep of w update}))}
 \nonumber\\
& \le \Lagr'_\rho(w_{t},z_{t+1},\lambda_{t}) +  
\t \langle w^* - w_{t}, \nabla_w \Lagr'_\rho(w_{t},z_{t+1},\lambda_{t}) \rangle
+  \frac{\t^2 \d_t^2}{2\a} \nonumber\\
& \qquad + \t R(w^*)+ (1-\t) R(w_t) + \sigma_{t+1} \|A_{t+1}\|_2^2 
\qquad (\text{convexity of } R  )\nonumber\\
& = \Lagr'_\rho(w_{t},z_{t+1},\lambda_{t})  +  
 \t \langle w^* - w_{t}, \nabla_w \Lagr'_\rho(w_{t},z_{t},\lambda_{t}) \rangle
+  \frac{\t^2\d_t^2}{2\a}\nonumber\\
& \qquad 
+ \t \langle w^* - w_t, \nabla_w \Lagr'_\rho(w_t,z_{t+1},\lambda_t) - \nabla_w \Lagr'_\rho(w_t,z_t,\lambda_t) \rangle \nonumber\\
& \qquad + \t R(w^*)+ (1-\t) R(w_t) +  \sigma_{t+1} \|A_{t+1}\|_2^2 . 
\label{eq:proof 1}
\end{align}
The last inner product above can be controlled as 
\begin{align}
& \t \langle w^* - w_t , \nabla_w \Lagr'_\rho(w_t,z_{t+1},\lambda_t) - \nabla_w \Lagr'_\rho(w_t,z_{t},\lambda_t) \rangle  \nonumber\\
& \le \frac{\t^2 \d_t^2}{2\a} + \frac{\a}{2}\| \nabla_w \Lagr'_\rho(w_t,z_{t+1},\lambda_t) - \nabla_w \Lagr'_\rho(w_t,z_{t},\lambda_t)  \|_2^2 
\qquad ( 2\langle a, b\rangle \le \|a\|_2^2+\|b\|_2^2 \text{ and (\ref{eq:defn of delta})})
\nonumber\\
& \le \frac{\t^2\d_t^2}{2\a} +  \a \tau^2_\rho \|z_{t+1}-z_t\|_2^2,
\qquad \text{(see (\ref{eq:smoothness lemma zw}))}
\end{align}
which, after substituting in \eqref{eq:proof 1}, yields that 
\begin{align}
& \Lagr'_\rho(v_{t+1},\lambda_{t+1}) + R(w_{t+1})  \nonumber\\
& \le \Lagr'_\rho(w_{t},z_{t+1},\lambda_{t}) +  
 \t \langle w^* - w_{t}, \nabla_w \Lagr'_\rho(w_{t},z_{t},\lambda_{t}) \rangle
+  \frac{\t^2\d_t^2}{\a} \nonumber\\
& \qquad + \a\tau_\rho^2 \|z_{t+1}-z_t\|_2^2  + \t R(w^*) + (1-\t) R(w_t)+  \sigma_{t+1} \|A_{t+1}\|_2^2. 
\label{eq:proof 1.1}
\end{align}
Regarding the right-hand side above, the smoothness of $\Lagr'_\rho$ in Lemma \ref{lem:smoothness} allows us to write that 
\begin{align}
& \Lagr'_\rho(w_t,z_{t+1},\lambda_t)
+ \a\tau_\rho^2 \|z_{t+1}-z_t\|_2^2 
\nonumber\\
& \le \Lagr'_\rho(w_t,z_t,\lambda_t) + 
\langle z_{t+1}-z_t, \nabla_z \Lagr'_\rho(w_t,z_t,\lambda_t)\rangle \nonumber\\
& \qquad + 
\l( \frac{\xi_\rho}{2} + \a\tau_\rho^2  \r)
 \|z_{t+1}-z_t\|_2^2.
\qquad \text{(see (\ref{eq:smoothness lemma z}))} 
\label{eq:proof 1.2}
\end{align}
If we assume that the primal step sizes $\alpha,\b$ satisfy
\begin{align}
\frac{\xi_\rho}{2}+ \a\tau_\rho^2 \le \frac{1}{2\b},
\label{eq:step size assump}
\end{align}
we can simplify \eqref{eq:proof 1.2} as 
\begin{align}
& \Lagr'_\rho(w_t,z_{t+1},\lambda_t) + \a\tau_\rho^2 \|z_{t+1}-z_t\|_2^2  \nonumber\\
& \le \Lagr'_\rho(w_t,z_t,\lambda_t) + 
\langle z_{t+1}-z_t, \nabla_z \Lagr'_\rho(w_t,z_t,\lambda_t)\rangle +
\frac{1}{2\b} \|z_{t+1}-z_t\|_2^2 . 
\qquad \text{(see (\ref{eq:step size assump}))}
\label{eq:proof 1.3}
\end{align}
From Algorithm~\ref{alg:admm}, recall the equivalent expression of the $z$ updates as 
\begin{align}
z_{t+1} = \arg\min_z \langle z - z_t , \nabla_z \Lagr'_\rho(w_t,z_t,\lambda_t) \rangle + \frac{1}{2\b}\|z-z_t\|_2^2 + H(z),
\label{eq:equiv_rep_of_z_updates}
\end{align}
and consider the choice of $z=\t z^* + (1-\t) z_t$ above, with $\t\in[0,1]$ to be set later.   Combining (\ref{eq:proof 1.3},\ref{eq:equiv_rep_of_z_updates}) leads us to 
\begin{align}
& \Lagr'_\rho(w_t,z_{t+1},\lambda_t ) + \a \tau_\rho^2 \|z_{t+1}-z_t\|_2^2 + H(z_{t+1})  \nonumber\\
& = \Lagr'_\rho(w_t,z_t,\lambda_t) + \min_z
\langle z - z_t, \nabla_z \Lagr'_\rho(w_t,z_t,\lambda_t) \rangle + 
\frac{1}{2\b}\|z-z_t\|_2^2
+ H(z) 
\qquad \text{(see (\ref{eq:proof 1.3},\ref{eq:equiv_rep_of_z_updates}))}
\nonumber\\
& \le 
\Lagr'_\rho(w_t,z_t,\lambda_t) + 
\t \langle z^* - z_t, \nabla_z \Lagr'_\rho(w_t,z_t,\lambda_t) \rangle + 
\frac{\t^2\d_t^{'2}}{2\b}
+ H(\t z^* + (1-\t) z_{t})
\nonumber\\
& 
\le \Lagr'_\rho(w_t,z_t,\lambda_t) + 
\t \langle z^* - z_t, \nabla_z \Lagr'_\rho(w_t,z_t,\lambda_t) \rangle + 
\frac{\t^2 \d_t^{'2}}{2\b} \nonumber\\
& \qquad + \t H(z^*) + (1-\t) H(z_t) .
\qquad (\text{convexity of } H ) 
\label{eq:proof 1.4}
\end{align} 
By combining (\ref{eq:proof 1.1},\ref{eq:proof 1.4}), we reach 
\begin{align}
& \Lagr_\rho(v_{t+1},\lambda_{t+1}) \nonumber\\
& = \Lagr'_\rho(v_{t+1},\lambda_{t+1}) + R(w_{t+1}) + H(z_{t+1}) 
\qquad \text{(see (\ref{eq:ind Lagr},\ref{eq:ind Lagr prime}))}
\nonumber\\
& \le \Lagr'_\rho(w_t,z_{t+1},\lambda_t) +
\t \langle w^* - w_t, \nabla_w \Lagr'_\rho(w_t,z_t,\lambda_t) \rangle + \frac{\t^2\d_t^2}{\a} 
+ \a\tau_\rho^2\|z_{t+1}-z_t\|_2^2 \nonumber\\
& \qquad + \t R(w^*) +(1-\t) R(w_t) + H(z_{t+1}) + \s_{t+1}\|A_{t+1}\|_2^2  
\qquad 
\text{(see (\ref{eq:proof 1.1}))}  \nonumber\\
& \le \Lagr'_\rho(v_t,\lambda_t)+ \t \langle v^* - v_t, \nabla_z \Lagr'_\rho(v_t,\lambda_t) \rangle  + \frac{\t^2\d_t^2}{\a}  + \frac{\t^2\d_t^{'2}}{2\b} \nonumber\\
& \qquad  + \t R(z^*)+ (1-\t) R(z_t) + \t H(z^*) + (1-\t) H(z_t) \nonumber\\
& \qquad + \s_{t+1}\|A_{t+1}\|_2^2 
\qquad \text{(see (\ref{eq:proof 1.4}))} 
 \nonumber\\
& = \Lagr_\rho(v_t,\lambda_t) + \t \langle v^* - v_t, \nabla_z \Lagr'_\rho(v_t,\lambda_t) \rangle + \frac{\t^2\d_t^2}{\a}+  \frac{\t^2\d_t^{'2}}{2\b} 
\nonumber\\
&\qquad + \t (R(z^*) + H(z^*)- R(z_t) - H(z_t)  ) 
+ \s_{t+1}\|A_{t+1}\|_2^2 
\qquad \text{(see (\ref{eq:ind Lagr},\ref{eq:ind Lagr prime}))} \nonumber\\
& \le \Lagr_\rho(v_t,\lambda_t) + \t \l( \frac{\omega_\rho \d_t^2}{2}+ \frac{\omega'_\rho \d_t^{'2}}{2}-\Delta'_t  \r)  + \frac{\t^2 \d_t^2}{\a} + \frac{\t^2 \d_t^{'2}}{2\b}
\nonumber\\
& \qquad  
+ \t (R(z^*) + H(z^*)- R(z_t) - H(z_t)  ) 
+ \s_{t+1}\|A_{t+1}\|_2^2 
\qquad \text{(see \eqref{eq:smoothness lemma 2})}
\nonumber\\
& = \Lagr_\rho(v_t,\lambda_t) + \t \l( \frac{\omega_\rho \d_t^2}{2}+ \frac{\omega'_\rho \d_t^{'2}}{2}-\Delta_t  \r) + \frac{\t^2 \d_t^{2}}{\a} + \frac{\t^2  \d_t^{'2}}{2\b}  \nonumber\\
& \qquad + \s_{t+1}\|A_{t+1}\|_2^2 
\qquad \text{(see (\ref{eq:ind Lagr},\ref{eq:ind Lagr prime}))}
\label{eq:proof 1.45}
\end{align}
%where we conveniently defined the shorthand 
%\begin{align}
%\g := \min(\b,\a). 
%\end{align}
After recalling (\ref{eq:defn of Delta}) and by subtracting $\Lagr_\rho(v^*,\lambda^*)$ from both sides,  \eqref{eq:proof 1.45} immediately implies that 
\begin{align}
 \Delta_{t+1}  & \le \Delta_t 
 + \frac{\omega_\rho \d_t^2}{2}+ \frac{\omega'_\rho \d_t^{'2}}{2}+ 
 \t \l(  \ol{\omega}_\rho-\Delta_t  \r) 
+ \frac{\t^2 \d_t^2 }{\a} + \frac{\t^2 \d_t^{'2} }{2\b} \nonumber\\
& \qquad  + \s_{t+1}\|A_{t+1}\|_2^2 ,
 \quad \text{(see (\ref{eq:defn of Delta},\ref{eq:proof 1.45}))}
\label{eq:proof 1.5}
\end{align}
where we also used the assumption that $\t\le 1$ above. 
%Assuming that 
%\begin{align}
%\b < \frac{1}{\xi_\rho},
%\label{eq:step size z cnd}
%\end{align}
%Let us set 
%\begin{align}
%\t_0 = \sqrt{\frac{2\a\b\tau_\rho^2}{1-\b\xi_\rho}},
%\end{align}
%assuming that $\b < 1/\xi_\rho$. 
To remove the feasibility gap $\|A_{t+1}\|_2$ from the right-hand side above, we write that 
\begin{align}
\|A_{t+1}\|_2 & = \| w_{t+1}-G(z_{t+1})\|_2 
\qquad \text{(see \eqref{eq:defn of A})}
\nonumber\\
& = \| w_{t+1}- w^* - (G(z_{t+1}) - G(z^*)) \|_2 
\qquad ( (w^*,z^*) \text{ is a solution of \eqref{eq:emp loss min}} )
\nonumber\\
& \le  \|w_{t+1}-w^*\|_2 + \| G(z_{t+1}) - G(z^*)\|_2 
\qquad \text{(triangle inequality)}
\nonumber\\
& \le  \|w_{t+1}-w^*\|_2 + \kappa_G \| z_{t+1} - z^*\|_2 
\qquad \text{(see \eqref{eq:lipschitz of G 2})}
\nonumber\\ 
& = \d_{t+1} + \k_G \d'_{t+1},
\qquad \text{(see \eqref{eq:defn of delta})}
\label{eq:bnd-on-At}
\end{align}
which, after substituting in \eqref{eq:proof 1.5}, yields that 
\begin{align}
\Delta_{t+1}  & \le \Delta_t 
 + \frac{\omega_\rho \d_t^2}{2}+ \frac{\omega'_\rho \d_t^{'2}}{2}+ 
 \t \l(  \ol{\omega}_\rho-\Delta_t  \r) 
+ \frac{\t^2 \d_t^2 }{\a} + \frac{\t^2 \d_t^{'2} }{2\b}+ 2\s_{t+1}\d_{t+1}^2  + 2\s_{t+1}\k_G^2 \d'^2_{t+1} \nonumber\\
& \qquad  
\qquad (\text{see \eqref{eq:bnd-on-At} and } (a+b)^2\le 2a^2+2b^2)  \nonumber\\
& \le \Delta_t 
 + \frac{\omega_\rho \d_t^2}{2}+ \frac{\omega'_\rho \d_t^{'2}}{2}+ 
 \t \l(  \ol{\omega}_\rho-\Delta_t  \r) 
+ \frac{\t^2 \d_t^2 }{\a} + \frac{\t^2 \d_t^{'2} }{2\b} + 2\s_0 \d_{t+1}^2  + 2\s_0 \k_G^2 \d'^2_{t+1}. \nonumber\\
& \qquad  .
\qquad (\s_{t+1} \le \s_0 \text{ in Algorithm \ref{alg:admm}})
\label{eq:bnd-for-all-thetas}
\end{align}
For  every iteration $t$, suppose that 
\begin{align}
\frac{\d_t^2}{\a} + \frac{\d_t^{'2}}{\b}  \ge \ol{\eta}_\rho \ge 
\frac{\ol{\mu}_\rho  }{ \min\l( \frac{\a \mu_\rho }{4}, \frac{\b\mu'_\rho}{2}  \r) - \sqrt{\max\l(\frac{\a}{2}(\omega_\rho+4\s_0), \b (\omega'_\rho+4\s_0 \k_G^2)\r)}}
,
\label{eq:thresh}
\end{align}
for $\ol{\eta}_\rho$ to be set later. Consequently, it holds that 
\begin{align}
\frac{\Delta_t }{\frac{2\d_t^2}{\a}+ \frac{\d'^2_t }{\b}} & \ge \frac{\frac{\mu_\rho \d_t^2}{2}+ \frac{\mu'_\rho \d'^2_t}{2} - \ol{\mu}_\rho }{ \frac{2\d_t^2}{\a}+ \frac{\d'^2_t}{\b} }
\qquad \text{(see \eqref{eq:str cvx lemma 1})} \nonumber\\
& \ge \min\l( \frac{\a\mu_\rho}{4}, \frac{\b\mu'_\rho}{2}  \r) - \frac{\ol{\mu}_\rho }{\frac{2\d_t^2}{\a}+ \frac{\d'^2}{\b}} \nonumber\\
&\ge  \min\l( \frac{\a\mu_\rho}{4}, \frac{\b\mu'_\rho}{2}  \r) - \frac{\ol{\mu}_\rho }{\ol{\eta}_\rho} 
\qquad \text{(see \eqref{eq:thresh})} 
\nonumber\\
& \ge \sqrt{\max\l(\frac{\a}{2}\l(\omega_\rho+4\s_0 \r), \b (\omega'_\rho+4\s_0 \k_G^2)\r)}. 
\qquad \text{(see \eqref{eq:thresh})}
\label{eq:show-well-defn}
\end{align}
We now set 
\begin{align}
\widehat{\theta}_t := \min\l(  
\sqrt{
\frac{\Delta_t ^2}{\l( \frac{2\d_t^2}{\a} + \frac{\d'^2_t}{\b} \r)^2}
- \max\l( \frac{\a}{2}\l( \omega_\rho+4\s_0 \r), \b\l(\omega'_\rho+4\s_0 \k_G^2 \r)  \r)}
, 1
 \r),
\label{eq:thetahat}
\end{align}
which is well-defined, as verified in  \eqref{eq:show-well-defn}. 
From (\ref{eq:show-well-defn},\ref{eq:thetahat}), it also immediately follows that 
\begin{align}
%\widehat{\theta}_t \in [\t_0,1],
 \widehat{\theta}_t \in [0,1], 
\qquad \forall t, 
\label{eq:range_of_theta}
\end{align}
\begin{align}
\Delta_t \ge0, \qquad \forall t,
\label{eq:admissible}
\end{align} 
which we will use later on in the proof. 
%where we use the convention that $[\t_0,1]=\{1\}$ if $\t_0 >1$. 
Consider first the case where $\widehat{\t}_t <1$. To study  the choice of $\theta = \widehat{\theta}_t$ in \eqref{eq:proof 1.5}, we will need the bound 
\begin{align}
& -\widehat{\t}_t\Delta_t + \widehat{\t}_t^2 \l( \frac{\d_t^2}{\a}+ \frac{\d'^2_t}{2\b} \r)  \nonumber\\
& = - \sqrt{
\frac{\Delta_t ^4}{\l( \frac{2\d_t^2}{\a} + \frac{\d'^2_t}{\b} \r)^2}
- \Delta_t^2 \max\l( \frac{\a}{2}\l(\omega_\rho+4\s_0 \r), \b\l(\omega'_\rho+4\s_0 \k_G^2 \r)  \r)} \nonumber\\
& \qquad + \frac{ \Delta_t^2}{\frac{4\d_t^2}{\a}+ \frac{2\d'^2_t}{\b} } - \max\l(\frac{\a}{2}\l(\omega_\rho +4\s_0 \r), \b\l(\omega'_\rho+4\s_0 \k_G^2 \r)  \r) \l( \frac{\d_t^2}{\a}+ \frac{\d'^2_t}{2\b} \r) 
\qquad \text{(see \eqref{eq:admissible})}
\nonumber\\
& \le - \frac{\Delta_t^2}{\frac{4\d_t^2}{\a}+\frac{2\d'^2}{\b}} + \Delta_t  \sqrt{\max\l(\frac{\a}{2}\l(\omega_\rho+4\s_0 \r), \b\l(\omega'_\rho+4\s_0 \k_G^2 \r)  \r)} \nonumber\\
& \qquad - \max\l(\frac{\a}{2}\l(\omega_\rho+4\s_0 \r), \b\l(\omega'_\rho+4\s_0 \k_G^2 \r)  \r) \l( \frac{\d_t^2}{\a}+ \frac{\d'^2_t}{2\b} \r),
\label{eq:nec-calc}
\end{align}
where the inequality above uses $\sqrt{a-b}\ge \sqrt{a}-\sqrt{b}$. Substituting  \eqref{eq:nec-calc} back into \eqref{eq:bnd-for-all-thetas}, we reach 
\begin{align}
\Delta_{t+1} & \le \Delta_t - \frac{\Delta_t^2}{\frac{4\d_t^2}{\a}+\frac{2\d'^2}{\b}} + \Delta_t  \sqrt{\max\l(\frac{\a}{2}\l(\omega_\rho+4\s_0 \r), \b\l(\omega'_\rho+4\s_0 \k_G^2 \r)  \r)} 
\qquad \text{(see (\ref{eq:bnd-for-all-thetas},\ref{eq:nec-calc}))}
\nonumber\\
& \le \Delta_t - \l( \min\l( \frac{\a\mu_\rho}{4}, \frac{\b\mu'_\rho}{2 } \r) - \frac{\ol{\mu}_\rho}{\ol{\eta}_\rho}  \r) \frac{\Delta_t }{2} \nonumber\\
& \qquad + \Delta_t  \sqrt{\max\l(\frac{\a}{2}\l(\omega_\rho+4\s_0 \r), \b\l(\omega'_\rho+4\s_0 \k_G^2 \r)  \r)} 
\qquad \text{(see third line of \eqref{eq:show-well-defn} and (\ref{eq:admissible}))}
\nonumber\\
& \le  \l(1 - \min\l( \frac{\a\mu_\rho}{8}, \frac{\b\mu'_\rho}{4 } \r) + \frac{\ol{\mu}_\rho}{2\ol{\eta}_\rho}  +
\sqrt{\max\l(\frac{\a}{2}\l(\omega_\rho+4\s_0 \r), \b\l(\omega'_\rho+4\s_0 \k_G^2 \r)  \r)}
\r) \Delta_t  \nonumber\\
%& \qquad + \min\l( \frac{\a \mu_\rho}{8},\frac{\b\mu'_\rho}{4} \r) \ol{\omega}_\rho+\a \ol{\tau}^2_\rho + \ol{\nu}_\rho + \ol{\xi}_\rho  \nonumber\\
& =: \eta_{\rho,1} \Delta_t ,
\qquad 
\text{if } \Delta_t  < \frac{\d_t^2}{\a}+ \frac{\d'^2_t}{\b}. 
\label{eq:contraction}
\end{align}
Next consider the case where $\widehat{\theta}_t=1$. With the choice of $\theta = \widehat{\theta}_t=1$ in \eqref{eq:bnd-for-all-thetas}, we find that 
\begin{align}
\Delta_{t+1}  
& \le
 \l( \frac{\omega_\rho}{2} + \frac{1}{\a}+ \rho \r) \d_t^2 
+ \l( \frac{\omega'_\rho}{2} + \frac{1}{2\b}+ \rho \kappa_G^2 \r) \d'^2_t 
\qquad \text{(see \eqref{eq:bnd-for-all-thetas})}\nonumber\\
& \le \frac{1}{2} \l( 1+ \max\l(  \frac{\a}{2} (\omega_\rho +4\s_0)    , \b(\omega'_\rho+4\s_0 \kappa_G^2  ) \r) \r)\cdot  \l( \frac{2\d_t^2}{\a}+ \frac{\d'^2_t}{\b}  \r) \nonumber\\
& \le \frac{1}{2} \sqrt{ 1+ \max\l(  \frac{\a}{2} (\omega_\rho +4\s_0)    , \b(\omega'_\rho+4\s_0 \kappa_G^2  ) \r) } \Delta_t  
\qquad \text{(see \eqref{eq:thetahat})} \nonumber\\
& =: \eta_{\rho,2} \Delta_t ,
\qquad \text{if } \Delta_t \ge  \frac{\d_t^2}{\a}+ \frac{\d'^2_t}{\b}.
\label{eq:contraction 2} 
\end{align}
To simplify the above expressions, let us assume that
\begin{align}
\sqrt{\max\l(\frac{\a}{2}(\omega_\rho+4\s_0), \b(\omega'_\rho+4\s_0\kappa_G^2)  \r)} \le  \min\l(\frac{\a\mu_\rho}{16},\frac{\b\mu'_\rho}{8} \r) \le \frac{1}{2},
\label{eq:simp-assump-contraction}
\end{align}
from which it follows that 
\begin{align}
\max(\eta_{\rho,1},\eta_{\rho,2}) &\le 1-  \min\l(\frac{\a\mu_\rho}{16}, \frac{\b\mu'_\rho}{8}  \r)+  \frac{\ol{\mu}_\rho}{2\ol{\eta}_\rho} \nonumber\\
& \le 1-   \min\l(\frac{\a\mu_\rho}{32}, \frac{\b\mu'_\rho}{16}  \r) \nonumber\\
& =: 1-\eta_\rho \in [0,1),
\label{eq:defn-of-eta}
\end{align}
where the second line above holds if
\begin{align}
\ol{\eta}_\rho \ge \frac{\ol{\mu}_\rho}{ \min\l(\frac{\a\mu_\rho}{16},\frac{\b\mu'_\rho}{8}  \r)}.
\label{eq:second-req-for-etabar}
\end{align}
Then, by unfolding (\ref{eq:contraction},\ref{eq:contraction 2}), we reach
\begin{align}
\Delta_t & \le 
(1-\eta_\rho)^t \Delta_0 .
\label{eq:geo decay lagrangian}
\end{align}
Moreover, by combining (\ref{eq:str cvx lemma 1},\ref{eq:geo decay lagrangian}), we can bound the error, namely,
\begin{align} 
\frac{\d_t^2}{\a} + \frac{\d'^2_t}{\b} 
& \le \max(\alpha \mu_\rho, \beta \mu'_\rho ) \l( \mu_\rho \d_t^2+ \mu'_\rho \d'^2_t \r) \nonumber\\
& \le \mu_\rho \d_t^2 + \mu'_\rho \d'^2_t 
\qquad \text{(see (\ref{eq:first-step-size-assump},\ref{eq:step size assump}), Lemmas \ref{lem:smoothness} and \ref{lem:str cvx})}\nonumber\\
& \le 2(\Delta_t + \ol{\mu}_\rho) 
\qquad \text{(see \eqref{eq:str cvx lemma 1})}
\nonumber\\
& \le 
 2 (1-\eta_\rho)^t   \Delta_0  + 
\frac{2\ol{\mu}_\rho}{\eta_\rho }
%\qquad \text{(see first line of \eqref{eq:geo decay lagrangian})}
\nonumber\\
& \le 2 (1-\eta_\rho)^t \Delta_0 + \frac{2 \ol{\mu}_\rho}{ \min\l( \frac{\a\mu_\rho}{16},\frac{\b\mu'_\rho}{8} \r)}  
\qquad \text{(see (\ref{eq:defn-of-eta}))}
\nonumber\\
& =: 2 (1-\eta_\rho)^t \Delta_0 +\frac{\ol{\eta}_\rho}{\rho}.
\qquad \text{(this choice of }\ol{\eta}_\rho \text{ satisfies (\ref{eq:thresh},\ref{eq:second-req-for-etabar}))}.
\label{eq:geo decay dist}
\end{align}
It remains to bound the feasibility gap $\|A_t\|_2$, see \eqref{eq:defn of A}.
\begin{comment}
The naive way to do achieve this is via \eqref{eq:bnd-on-At}, which does not clearly reflect the role of penalty weight $\rho$. Indeed, increasing penalty weight $\rho$ must eventually derive  the feasibility gap to zero, which is not reflected by the right-hand side of  \eqref{eq:bnd-on-At}, as clear from  \eqref{eq:thresh}. 
\end{comment}
Instead of \eqref{eq:bnd-on-At}, we consider the following alternative approach to bound $\|A_t\|_2$. Using definition of $\Delta_t$ in  \eqref{eq:defn of Delta}, we write that 
\begin{align}
\Delta_t & = \mathcal{L}_\rho(v_t,\lambda_t) - \mathcal{L}_\rho(v^*,\lambda^* ) 
\qquad \text{(see \eqref{eq:defn of Delta})}
\nonumber\\
& = \mathcal{L}_\rho(v_t,\lambda_t) - \mathcal{L}_\rho(v_t,\lambda^*) + \mathcal{L}_\rho(v_t,\lambda^*) - 
\mathcal{L}_\rho(v^*,\lambda^* ) \nonumber\\
& = \langle A_t, \lambda_t - \lambda^* \rangle + \mathcal{L}(v_t,\lambda^*) - \mathcal{L}(v^*,\lambda^*) + \frac{\rho}{2}\|A_t\|_2^2 ,
\label{eq:Delta-to-A-1}
\end{align}
where 
\begin{align}
\mathcal{L}(v,\lambda)=\mathcal{L}(w,z,\lambda) := L(w) + R(w) + H(z) + \langle w- G(z), \lambda\rangle.
\end{align}
It is not difficult to verify that $\mathcal{L}(v^*,\lambda^*)=\mathcal{L}_\rho(v^*,\lambda^*)$ is the optimal value of problem \eqref{eq:emp loss min} and that $\mathcal{L}(v_t,\lambda^*)\ge \mathcal{L}(v^*,\lambda^*)$, from which it follows that 
\begin{align}
\Delta_t & \ge \langle A_t, \lambda_t - \lambda^* \rangle + \frac{\rho}{2}\|A_t\|_2^2 
\qquad \text{(see \eqref{eq:Delta-to-A-1})}
\nonumber\\
& \ge - \frac{\rho}{4}\|A_t\|_2^2- \frac{1}{\rho }\|\lambda_t - \lambda^*\|_2^2 + \frac{\rho}{2}\|A_t\|_2^2 
\qquad (\text{Holder's inequality and } 2ab\le a^2+b^2) \nonumber\\
& \ge - \frac{2}{\rho}\|\lambda_t\|_2^2 - \frac{2}{\rho}\|\lambda^*\|_2^2 + \frac{\rho}{4}\|A_t\|_2^2
\qquad ((a+b)^2 \le 2 a^2+2b^2) \nonumber\\
& \ge -\frac{2\lambda_{\max}^2}{\rho} - \frac{2\|\lambda^*\|_2^2}{\rho}  + \frac{\rho}{4}\|A_t\|_2^2,
\qquad \text{(see \eqref{eq:dual seq})}
\label{eq:Delta-to-A-2}
\end{align}
which, in turn, implies that 
\begin{align}
\|A_t\|_2^2 &\le \frac{4}{\rho}\l(\Delta_t +\frac{2\lambda_{\max}^2}{\rho} + \frac{2\|\lambda^*\|_2^2}{\rho}\r)
\qquad \text{(see \eqref{eq:Delta-to-A-2})} \nonumber\\
& \le \frac{4}{\rho}\l( (1-\eta_\rho)^t \Delta_0 + \frac{2(\ol{\eta}_{\rho,1}+ \ol{\eta}_{\rho,2})}{\eta_\rho}+\frac{2\lambda_{\max}^2}{\rho} + \frac{2\|\lambda^*\|_2^2}{\rho} \r)
\qquad \text{(see \eqref{eq:geo decay lagrangian})} \nonumber\\
& \le \frac{4}{\rho}\l( (1-\eta_\rho)^t \Delta_0 + \frac{\ol{\eta}_\rho +2\lambda_{\max}^2 + 2\|\lambda^*\|_2^2}{\rho} \r)
\qquad \text{(see \eqref{eq:geo decay dist})}\nonumber\\
& =: \frac{4(1-\eta_\rho)^t \Delta_0}{\rho }+ \frac{\widetilde{\eta}_\rho}{\rho^2}.
\label{eq:bnd on feas proof}
%\frac{4}{\rho}\l(\ol{\eta}_\rho + 2\lambda_{\max}^2+ 2\|\lambda^*\|_2^2\r) 
\end{align}
This completes the proof of Theorem \ref{thm:main}.

Let us also inspect the special case  where $\mu_L\gg \rho \gtrsim 1$ and $\iota_G^2 \gg \nu_G$, where $\approx$ and $\gtrsim$ suppress any universal constants and dependence on the dual optimal variable $\lambda^*$, for the sake of simplicity.   From Lemmas \ref{lem:smoothness} and \ref{lem:str cvx}, it is easy to verify that 
\begin{align*}
\nu_\rho \approx \nu_L,
\qquad \xi_\rho \approx \rho \kappa_G^2,
\qquad 
\tau_\rho = \rho \kappa_G,
\end{align*}
\begin{align}
\mu_\rho \approx \mu_L,
\qquad 
\mu'_\rho \approx \rho \iota_G^2,
\qquad 
\ol{\mu}_\rho \approx \rho^{-1},
\qquad 
\omega'_\rho \approx \rho \nu_G.
\end{align}
%\begin{align}
%\ol{\nu}_\rho = \ol{\xi}_\rho=\ol{\tau}_\rho = \omega_\rho = \ol{\omega}_\rho = 0.
%\end{align}
We can then take 
\begin{align}
\alpha & \approx \frac{1}{\nu_L}, \qquad \text{(see \eqref{eq:first-step-size-assump})} \nonumber\\
\beta & \approx \frac{1}{\xi_\rho} \approx \frac{1}{\rho \kappa_G^2},
\qquad \text{(see \eqref{eq:step size assump})} \nonumber\\
\eta_\rho &  \approx \min\l(\frac{\mu_L}{\nu_L}, \frac{\iota_G^2}{\kappa_G^2}  \r),
\qquad \text{(see \eqref{eq:defn-of-eta})}\nonumber\\
%\ol{\eta}_{\rho,1} &= \ol{\eta}_{\rho,2}  = 0,
%\qquad \text{(see (\ref{eq:contraction},\ref{eq:contraction 2}))}
%\nonumber\\
\ol{\eta}_\rho & \approx \frac{\rho \ol{\mu}_\rho}{\min\l(\a\mu_\rho, \b\mu'_\rho \r)} \approx \max\l( \frac{\nu_L}{\mu_L}, \frac{\kappa_G^2}{\iota_G^2}  \r),
\qquad \text{(see \eqref{eq:geo decay dist})} \nonumber\\
\widetilde{\eta}_\rho & \approx \ol{\eta}_\rho \approx   \max\l( \frac{\nu_L}{\mu_L}, \frac{\kappa_G^2}{\iota_G^2}  \r).
\qquad \text{(see \eqref{eq:bnd on feas proof})}
\end{align}
Lastly, for \eqref{eq:simp-assump-contraction} to hold, it suffices that 
\begin{align}
\s_0 \lesssim \rho  \min\l(\frac{\mu_L^2}{\nu_L^2}, \frac{\iota_G^4}{\kappa_G^4}  \r)=:\s_{0,\rho}.
\end{align}

\section{Proof of Lemma \ref{lem:smoothness} \label{sec:smoothness}}

To prove \eqref{eq:smoothness lemma w}, we write that 
\begin{align}
&  \Lagr'_\rho (w_{t+1},z_{t+1},\lambda_{t}) - \Lagr'_\rho (w_{t},z_{t+1},\lambda_{t}) - 
\langle w_{t+1}-w_{t}, \nabla_w \Lagr'_\rho(w_{t},z_{t+1},\lambda_{t})  \rangle  \nonumber\\
& = L(w_{t+1}) - L(w_t) - \langle w_{t+1}-w_t , \nabla_w L(w_t) \rangle \nonumber\\
& \qquad+ \frac{\rho}{2}\|w_{t+1}-G(z_{t+1})\|_2^2 - \frac{\rho}{2}\|w_t - G(z_{t+1})\|_2^2 - 2\rho \langle w_{t+1}-w_t, w_t - G(z_{t+1}) \rangle  
\qquad \text{(see \eqref{eq:ind Lagr prime})}
\nonumber\\
& \le \frac{\nu_L}{2}\|w_{t+1}-w_t\|_2^2 + \ol{\nu}_L + \frac{\rho}{2}\| w_{t+1}-w_t \|_2^2 
\qquad \text{(see \eqref{eq:smoothness of L})} \nonumber\\
& =: \frac{\nu_\rho}{2}\|w_{t+1}-w_t\|_2^2 + \ol{\nu}_\rho.
\end{align}
To prove \eqref{eq:smoothness lemma z}, let us first control the dual sequence $\{\lambda_{t}\}_t$ by writing that 
\begin{align}
\|\lambda_{t}\|_2 & = 
\| \lambda_{0}+ \sum_{i=1}^t \sigma_{i} A_{i} \|_2 
\qquad \text{(see \eqref{eq:dual update proof})}
\nonumber\\
& \le \| \lambda_{0}\|_2 + \sum_{i=1}^t \sigma_{i} \|A_{i}\|_2 
\qquad \text{(triangle inequality)}
\nonumber\\
& \le \| \lambda_{0}\|_2 + \sum_{t'=1}^t \frac{\s_0}{i \log^2 (i+1)} \nonumber\\
& \le   \| \lambda_{0}\|_2 + c\sigma_0 \nonumber\\
& =: \lambda_{\max},
\label{eq:dual seq}
\end{align}
where 
\begin{align}
c \ge \sum_{t=1}^{\infty} \frac{1}{t \log^2(t+1)}. 
\end{align}
We now write that 
\begin{align}
& \Lagr'_\rho(w_t,z_{t+1},\lambda_t) - \Lagr'_\rho(w_t,z_{t},\lambda_t) - \langle z_{t+1}-z_t, \nabla_z \Lagr'_\rho(w_t,z_t,\lambda_t) \nonumber\\
& = -\langle G(z_{t+1}) - G(z_t) - DG(z_t) (z_{t+1}-z_t) ,\lambda_t \rangle \nonumber\\
& \qquad + \frac{\rho}{2} \|w_t - G(z_{t+1})\|_2^2 - \frac{\rho}{2}\|w_t - G(z_t)\|_2^2 \nonumber\\
& \qquad + \rho \langle DG(z_t) (z_{t+1}-z_t) ,  w_t - G(z_t) \rangle .
\qquad \text{(see \eqref{eq:ind Lagr prime})}
\label{eq:lem proof 1}
\end{align}
To bound the first inner product on the right-hand side above, we write that 
\begin{align}
& \langle G(z_{t+1}) - G(z_t) - DG(z_t) (z_{t+1}-z_t) ,\lambda_t \rangle \nonumber\\
& \le \| G(z_{t+1}) - G(z_t) - DG(z_t) (z_{t+1}-z_t) \|_2 \cdot \|\lambda_t\|_2
\qquad \text{(Cauchy-Shwartz's inequality)} \nonumber\\
& \le \frac{\nu_G\lambda_{\max}}{2} \|z_{t+1}-z_t\|_2^2  
\qquad \text{(see (\ref{eq:smoothness of G},\ref{eq:dual seq}))}
\label{eq:lem proof 2}
\end{align}
The remaining component on the right-hand side of \eqref{eq:lem proof 1} can be bounded as 
\begin{align}
& \|w_t-G(z_{t+1})\|_2^2 - \|w_t - G(z_t)\|_2^2 +2  \langle DG(z_t)(z_{t+1}-z_t), w_t - G(z_t) \rangle \nonumber\\
& = \|w_t-G(z_{t+1})\|_2^2 - \|w_t - G(z_t)\|_2^2  
+ 2 \langle G(z_{t+1}) - G(z_t), w_t- G(z_t) \rangle 
\nonumber\\
& \qquad -2 \langle G(z_{t+1}) - G(z_t)  -DG(z_t) (z_{t+1}-z_t) , w_t -G(z_t) \rangle  \nonumber\\
& = \|G(z_{t+1})-G(z_t) \|_2^2 
\nonumber\\
& \qquad +2 \langle G(z_{t+1}) - G(z_t) - DG(z_t) (z_{t+1}-z_t) , w_t -G(z_t) \rangle \nonumber\\
& \le  \|G(z_{t+1})-G(z_t) \|_2^2 
\nonumber\\
& \qquad +2  \|G(z_{t+1}) - G(z_t) - DG(z_t) (z_{t+1}-z_t) \|_2 \cdot \| w_t -G(z_t)\|_2
\qquad \text{(Cauchy-Shwartz's inequality)} \nonumber\\
& \le  \kappa_G^2 \|z_{t+1}-z_t\|_2^2 
+  \nu_G\| z_{t+1}-z_t \|_2^2   \|w_t - G(z_t) \|_2
\qquad 
\text{(see (\ref{eq:smoothness of G},\ref{eq:lipschitz of G 2}))} \nonumber\\
& \le   \kappa_G^2 \|z_{t+1}-z_t\|_2^2 + \nu_G\|z_{t+1}-z_t\|_2^2 \max_i  \|A_i\|_2. 
\qquad \text{(see (\ref{eq:defn of A}))}
\label{eq:lem proof 3}
\end{align}
Substituting the bounds in (\ref{eq:lem proof 2},\ref{eq:lem proof 3}) back into \eqref{eq:lem proof 1}, we find that 
\begin{align}
& \Lagr'_\rho(w_t,z_{t+1},\lambda_t) - \Lagr'_\rho(w_t,z_{t},\lambda_t) - \langle z_{t+1}-z_t, \nabla_z \Lagr'_\rho(w_t,z_t,\lambda_t) \nonumber\\
& \le 
\frac{1}{2}\l(   \nu_G (\lambda_{\max}+\rho \max_i \|A_i\|_2 ) + \rho \kappa_G^2\r) \|z_{t+1}-z_t\|_2^2 \nonumber\\
& =: \frac{ \xi_\rho}{2}\|z_{t+1}-z_t\|_2^2 + \ol{\xi}_\rho, 
\end{align}
which proves \eqref{eq:smoothness lemma z}. To prove \eqref{eq:smoothness lemma zw}, we write that 
\begin{align}
& \| \nabla_w \Lagr'_\rho(w_t,z_{t+1},\lambda_t) - \nabla_w \Lagr'_\rho(w_t,z_t,\lambda_t) \|_2 \nonumber\\
& = \rho \| G(z_{t+1}) - G(z_t) \|_2  
\qquad \text{(see \eqref{eq:ind Lagr prime})} \nonumber\\
& \le \rho  \kappa_G \|z_{t+1}-z_t\|_2 
\qquad \text{(see \eqref{eq:lipschitz of G 2})} \nonumber\\
& =: \tau_\rho \| z_{t+1}-z_t\|_2 + \ol{\tau}_\rho. 
\end{align}
This completes the proof of Lemma \ref{lem:smoothness}. 

\section{Proof of Lemma \ref{lem:str cvx} \label{sec:str cvx}}

For future reference, we record that 
\begin{align}
&   \langle v_{t} - \wh{v}, \nabla_v \Lagr'_\rho (\wh{v}) \rangle 
\nonumber\\
& = \langle w_{t} - \wh{w} , \nabla_w \Lagr'_\rho (\wh{v}) \rangle 
+ \langle z_{t} -\wh{z} , \nabla_z \Lagr'_\rho (\wh{v})  \rangle 
\qquad \l( v = (w,z) \r)
\nonumber\\
& = \langle w_{t} - \wh{w} , \nabla L(\wh{w}) + \wh{\lambda} + \rho (\wh{w}-G(\wh{z}) \rangle 
- \langle DG(\wh{z})(z_{t} - \wh{z}) , \wh{\lambda} + \rho (\wh{w} - G(\wh{z})) \rangle 
\qquad \text{(see \eqref{eq:ind Lagr prime})}
 \nonumber\\
 & = \langle w_{t} - \wh{w} , \nabla L(\wh{w}) + \wh{\lambda}  \rangle 
- \langle DG(\wh{z})(z_{t} - \wh{z}) , \wh{\lambda} \rangle,
\qquad \qquad \text{(see \eqref{eq:opt cnds})}
\label{eq:opt cnd consq}
\end{align}
where the last line above uses the feasibility of $v^*$ in \eqref{eq:emp loss min}. 
To prove \eqref{eq:str cvx lemma 1}, we use the definition of $\Lagr_\rho$ in \eqref{eq:ind Lagr} to write that 
\begin{align}
& \Lagr_\rho(v_{t},\lambda_{t}) - \Lagr_\rho(\wh{v},\wh{\lambda}) \nonumber\\
& = \Lagr'_\rho(v_t,\lambda_t) - \Lagr'_\rho(v^*,\lambda^*) 
+ R(w_t) - R(w^*) + L(z_t) - L(z^*) 
\qquad \text{(see (\ref{eq:ind Lagr},\ref{eq:ind Lagr prime}))}
\nonumber\\
& \ge \Lagr'_\rho(v_t,\lambda_t) - \Lagr'_\rho(v^*,\lambda^*)  
- \langle v_t - v^* ,
\nabla_v \Lagr'_\rho (v^*,\lambda^* ) 
 \rangle 
 \qquad \text{(see (\ref{eq:opt cnds}))}
  \nonumber\\
  & = L(w_{t}) - L(\wh{w}) - \langle w_{t} - \wh{w}, \nabla L(\wh{u}) \rangle \nonumber\\
& \qquad + \langle A_{t} , \lambda_{t} \rangle - \langle w_{t} - \wh{w}  - DG(\wh{z}) (z_{t} - \wh{z}),  \wh{\lambda} \rangle + \frac{\rho}{2}\|A_{t}\|_2^2 
\qquad \text{(see (\ref{eq:opt cnd consq}))} 
\nonumber\\
& \ge \frac{\mu_L \d_t^2}{2} + \langle A_{t}, \lambda_{t}- \wh{\lambda}\rangle + \frac{\rho}{2}\|A_{t}\|_2^2 \nonumber\\
& \qquad + \langle G(z_{t}) - G(\wh{z}) - DG(\wh{z}) (z_{t} - \wh{z}_k) , \wh{\lambda} \rangle 
\qquad \text{(see (\ref{eq:str cvx of L},\ref{eq:defn of delta}))} 
\nonumber\\
& \ge \frac{\mu_L\d_t^2}{2} + \langle A_{t}, \lambda_{t}- \wh{\lambda}\rangle + \frac{\rho}{2}\|A_{t}\|_2^2 -\frac{\nu_G\d'^2_t}{2}      \| \lambda^*\|_2. 
\qquad \text{(see (\ref{eq:smoothness of G},\ref{eq:defn of delta}))}
\label{eq:lwr bnd 0}
\end{align}
To control the terms involving $A_t$ in the last line above, we write that 
\begin{align}
& \langle A_t, \lambda_t - \lambda^* \rangle + \frac{\rho}{2}\|A_t\|_2^2 \nonumber\\
& = \frac{\rho}{2}\l\| A_t - \frac{\lambda_t - \lambda^*}{\rho}\r\|_2^2 - \frac{\|\lambda_t-\lambda^*\|_2^2}{2\rho} \nonumber\\
& = \frac{\rho}{2}\l\| w_t - w^* - (G(z_t) - G(z^*)) - \frac{\lambda_t - \lambda^*}{\rho}  \r\|_2^2  - \frac{\|\lambda_t-\lambda^*\|_2^2}{2\rho}
\qquad \text{(see (\ref{eq:opt cnds},\ref{eq:defn of A}))} \nonumber\\
& \ge \frac{\rho}{4}\|G(z_t)-G(z^*)\|_2^2 - \rho\d_t^2 - \frac{3\|\lambda_t - \lambda^*\|_2^2}{2\rho}
\qquad \l(  \|a-b-c\|_2^2 \ge \frac{\|a\|_2^2}{2} - 2\|b\|_2^2 - 2\|c\|_2^2  \r) \nonumber\\
& \ge \frac{\rho\iota_G^2\d'^2_t}{4}  - \rho\d_t^2 - \frac{3\|\lambda_t - \lambda^*\|_2^2}{2\rho}
\qquad \text{(see (\ref{eq:defn of delta},\ref{eq:lipschitz of G 2}))} \nonumber\\
& \ge \frac{\rho \iota_G^2 \d'^2_t}{4}  - \rho\d_t^2 - \frac{3  }{\rho}(\lambda_{\max}^2 + \|\lambda^*\|_2^2),
\qquad ( (a+b)^2 \le 2a^2 +2b^2 \text{ and (\ref{eq:dual seq})})
\end{align}
which, after substituting in \eqref{eq:lwr bnd 0}, yields that 
\begin{align}
& \Lagr_\rho(v_t,\lambda_t) - \Lagr_\rho(v^*,\lambda^*)  \nonumber\\
& \ge \frac{\mu_L - 2\rho }{2}\d_t^2 + \frac{1}{2}\l( \frac{\rho\iota^2_G}{2} - \nu_G\|\lambda^*\|_2 \r) \d'^2_t  - 
\frac{3}{\rho}\l( \lambda_{\max}^2+ \|\lambda^*\|_2^2\r) \nonumber\\
& \ge \frac{\mu_\rho \d_t^2}{2}+ \frac{\mu_\rho'\d'^2_t}{2} - \ol{\mu}_\rho,
\end{align}
where 
\begin{align}
\mu_\rho :=  \mu_L - 2\rho,
\qquad 
\mu'_\rho := \frac{\rho {\iota}^2_G}{2}-\nu_G\|\lambda^*\|_2 ,
\end{align}
\begin{align}
\ol{\mu}_\rho :=  
\frac{3}{\rho}\l( \lambda_{\max}^2 + \|\lambda^*\|_2^2\r). 
\end{align}
This proves \eqref{eq:str cvx lemma 1}. To prove \eqref{eq:smoothness lemma 2}, we use the definition of $\Lagr'_\rho$ in \eqref{eq:ind Lagr prime} to write that 
\begin{align}
& \Lagr'_\rho(v^*,\lambda^*) - \Lagr'_\rho(v_{t},\lambda_{t}) - \langle v^*- v_{t}, \nabla_v \Lagr'_\rho(v_{t},\lambda_{t}) \rangle \nonumber\\
& = L(w^*) - L(w_{t}) - \langle w^* - w_{t} , \nabla L(w_{t}) \rangle \nonumber\\
& \qquad - \langle 
A_t + DA(v_t)(v^*- v_{t})  
,\lambda_{t} \rangle\nonumber\\
& \qquad 
- \frac{\rho}{2}\langle A_{t} + 2 DA(v_t) (v^* - v_{t}) , A_{t} \rangle,
\qquad \text{(see \eqref{eq:ind Lagr prime})}
\label{eq:str cvx lem 0}
\end{align}
where 
\begin{align}
D A(v) & = \l[
\begin{array}{cc}
I_d & - DG(z)
\end{array}
\r],
\label{eq:defn of Jacobian of A}
\end{align}
is the Jacobian of the map $A$. 
The second inner product on the right-hand side of \eqref{eq:str cvx lem 0} can be bounded as 
\begin{align}
&  - \langle 
A_t + DA(v_t)(v^*- v_{t})  
,\lambda_{t} \rangle  \nonumber\\
& =- \langle 
w_{t}-G(z_{t}) + (w^*- w_{t}) - DG(z_{t})(z^* - z_{t})
,\lambda_{t} \rangle 
\qquad \text{(see (\ref{eq:defn of A},\ref{eq:defn of Jacobian of A}))}
\nonumber\\
& = -\langle 
G(z^*) -G(z_{t}) -  DG(z_{t})(z^* - z_{t})
,\lambda_{t} \rangle 
\qquad \l( w^* = G(z^*) \r) \nonumber\\
& \ge- \frac{\nu_G\d'^2_t}{2}   \|\lambda_{t}\|_2   
\qquad \text{(see (\ref{eq:smoothness of G},\ref{eq:defn of delta}))} \nonumber\\
& \ge -\frac{\nu_G\d'^2_t}{2}  \lambda_{\max}.
\qquad \text{(see \eqref{eq:dual seq})} 
\label{eq:lwr bnd proof 0}
\end{align}
To control  the last  inner product on the right-hand side of  \eqref{eq:str cvx lem 0}, we write that 
\begin{align}
& - \frac{\rho}{2}\langle A_t + 2 DA(v_t) (v^* - v_t), A_t \rangle \nonumber\\
& = \frac{ \rho}{2} \|A_t\|_2^2 - \rho \langle A_t + DA(v_t)(v^* - v_t), A_t \rangle \nonumber\\
& \ge -\rho \|A_t +DA(v_t)(v^*-v_t)\|_2 \|A_t\|_2
\qquad \text{(Holder's inequality)}
 \nonumber\\
& = -\rho \| (w^* - G(z^*)) - (w_t - G(z_t))  - (w^* - w_t) + DG(z_t) (z^* - z_t) \|_2 
\qquad (\text{see (\ref{eq:defn of A},\ref{eq:defn of Jacobian of A}) and } w^* = G(z^*) )  
\nonumber\\
& = - \rho \| G(z^*) - G(z_t) - DG(z_t) (z^* - z_t) \|_2 \nonumber\\
& \ge - \frac{\rho \nu_G}{2}\|z^* - z_t\|_2^2 
\qquad \text{(see \eqref{eq:smoothness of G})}
\nonumber\\
&  = - \frac{\rho \nu_G \d'^2_t}{2}.
\qquad \text{(see \eqref{eq:defn of delta})}
\label{eq:lwr bnd proof 3}
\end{align}
By substituting the bounds in  (\ref{eq:lwr bnd proof 0},\ref{eq:lwr bnd proof 3}) back into (\ref{eq:str cvx lem 0}) and also using the convexity of $L$, we reach 
\begin{align}
&  \Lagr'_\rho(v^*,\lambda^*) - \Lagr'_\rho(v_{t},\lambda_{t}) - \langle v^*- v_{t}, \nabla_v \Lagr'_\rho(v_{t},\lambda_{t}) \rangle \nonumber\\
& \ge - \frac{\nu_G  }{2} \l(\lambda_{\max}+ \rho   \r)
 \d'^2_t. 
\end{align}
This proves \eqref{eq:smoothness lemma 2}, thus completing the proof of Lemma \ref{lem:str cvx}.

\section{Relation with Gradient Descent \label{sec:gd-admm}}

%This section uncovers the relation between several first-order algorithms that have been proposed in the literature to solve Program~\eqref{eq:emp loss min}. Loosely speaking, we will establish that Algorithm~\ref{alg:admm} is closely related to previous algorithms, and that our convergence analysis in Section~\ref{sec:guarantees1} also applies to the algorithm proposed in \cite{Bora2017} without rates. We caution the reader that the discussion in this section is less formal by its nature. 
%
%Let us now turn to the details. 
Throughout this section, we set $R\equiv 0$ and $H\equiv 0$ in problem~\eqref{eq:emp loss min} and consider the updates in Algorithm~2, namely, 
\begin{equation}
\begin{split}
    z_{ t+1} &=  z_{ t} - \b \grad_{z} \Lagr_{\rho} (w_{t}, z_{t},  \lambda_{t}) , \\
    w_{ t+1} &\in \underset{w}{\operatorname{argmin}} \,\, \Lagr_\rho(w,z_{t+1},\lambda_t), \\
    \lambda_{t+1} &= \lambda_{ t} + \sigma_{t+1}  (w_{t+1}-G(z_{t+1})).
\end{split}
\label{eq:aug lagr alg full}
\end{equation}
From \eqref{eq:defn_aug_lagr}, recall that $\Lagr_\rho(w,z,\lambda)$ is convex in $w$ and the second step in \eqref{eq:aug lagr alg full} is therefore often easy to implement  with any over-the-shelf standard convex solver. 
 Recalling~(\ref{eq:defn_aug_lagr}), note also that the optimality condition for $w_{t+1}$ in \eqref{eq:aug lagr alg full} is  
\begin{align}
w_{t+1} - G(z_{t}) & = -\frac{1}{\rho}(\nabla L_m(w_{t+1})+ \lambda_t).
\label{eq:feas_to_grads}
\end{align}
Using \eqref{eq:defn_aug_lagr} again, we also write that 
\begin{align}
& \nabla_z \Lagr_\rho(w_{t+1},z_t,\lambda_t) \nonumber\\
& = - DG(z_t)^\top ( \lambda_t +\rho (w_{t+1}-G(z_t)) \nonumber\\
& = - DG(z_t)^\top ( \lambda_t  - \lambda_{t-1}- \nabla L_m(w_t)  ) \nonumber\\
& = - DG(z_t)^\top  ( \s_t (w_t - G(z_t)) - \nabla L(w_t)  ) ,
\label{eq:nablaL}
\end{align}
where the last two lines above follow from (\ref{eq:feas_to_grads},\ref{eq:aug lagr alg full}), respectively. Substituting back into the $z$ update in \eqref{eq:aug lagr alg full}, we reach
\begin{align}
z_{t+1} & = z_t + \b \s_t DG(z_t)^\top (w_t - G(z_t)) - \b \nabla L(w_t) 
\qquad \text{(see (\ref{eq:aug lagr alg full},\ref{eq:nablaL}))},
\label{eq:collapsed-admm}
\end{align}
from which it follows that 
\begin{align}
& \l\|z_{t+1}-(z_t -\b \nabla L(G(z_t)) )   \r\|_2 \nonumber\\
& \le \b\s_t \| DG(z_t)^\top (w_t - G(z_t)) \|_2+ \b \| \nabla L(w_t) - \nabla L( G(z_t) ) \|_2
\qquad \text{(see \eqref{eq:collapsed-admm})} \nonumber\\
& \le  \b \l( \s_t \kappa_G+ \nu_L \r) \| w_t - G(z_t) \|_2 .
\qquad \text{(see Assumptions \ref{assump:risk} and \ref{defn:G_iso})}
\end{align}
That is, as the feasibility gap vanishes in \eqref{eq:feas-thm} in Theorem \ref{thm:main},  the updates of Algorithm 2 match those of GD.
%
%When $t\gg 1$, the choice of dual step sizes in Algorithm~\ref{alg:admm} satisfies $\s_t\|w_t - G(z_t)\|_2 \approx 0$. Therefore, the $z$ update in \eqref{eq:aug lagr alg full} simplifies to 
%\begin{align}
%z_{t+1} \approx z_t - \b DG(z_t)^\top \nabla L_m(w_t),
%\label{eq:z_approx}
%\end{align}
%provided that $G$ is sufficiently smooth. 
%In addition, when $\rho \gg 1$, the $w$ update in \eqref{eq:aug lagr alg full} ensures that $w_t \approx G(z_t)$ which, together with \eqref{eq:z_approx},  implies that 
%\begin{align}
%z_{t+1} \approx z_t - \b DG(z_t)^\top \nabla L_m(G(z_t)), 
%\label{eq:grad_step}
%\end{align}
%if the  prior $G$ and empirical loss $L_m$ are sufficiently smooth. Note that \eqref{eq:grad_step} corresponds to one step of gradient descent (GD) with step size $\b$ applied to the program 
%\begin{align}
%\min_z \,\,L_m(G(z)),
%\label{eq:parameter_space}
%\end{align}
%which is equivalent to Program~\eqref{eq:emp loss min} with $R=H=0$. That is, for a sufficiently large penalty weight, the updates in \eqref{eq:aug lagr alg full} approximately coincide  with those in GD applied to  Program~\eqref{eq:parameter_space}. Since the updates in \eqref{eq:aug lagr alg full} can only improve over Algorithm~\ref{alg:admm}, the convergence rate in Section~\ref{sec:guarantees1} also applies to GD for Program~\eqref{eq:parameter_space}, thereby providing convergence rate as a special case for the algorithm introduced in~\citet{Bora2017} without supporting theory. 

\section{Proof of Lemma \ref{lem:ws_to_wn} \label{sec:proof of ws_to_wn}}

Recall that $R=1_W$ and $H\equiv 0$ for this proof. Using the optimality of $w^*\in\text{relint}(W)$ in~\eqref{eq:emp loss min supp}, we can write that 
\begin{align}
\| \nabla L(w^*) \|_2 & \le \| \nabla L_m(w^*) \|_2 + \| \nabla L_m(w^*) - \nabla L(w^*) \|_2
\qquad 
\text{(triangle inequality)} \nonumber\\
& =  \| \nabla L_m(w^*) - \nabla L(w^*) \|_2
\qquad \l( \nabla L_m(w^*) = 0  \r) \nonumber\\
& \le \max_{w\in W}\| \nabla L_m(w) - \nabla L(w) \|_2.
\label{eq:near_st}
\end{align}
On the other hand, using the strong convexity of $L$ in \eqref{eq:L_str_cvx_supp}, we can write that 
\begin{align}
\| w^\n - w^* \|_2 &
\le \frac{1}{\zeta_L} \| \nabla L(w^\n) - \nabla L(w^*) \|_2 
\qquad \text{(see \eqref{eq:L_str_cvx_supp})}
\nonumber\\
& = \frac{1}{\zeta_L} \|\nabla L(w^*) \|  
\qquad (\nabla L(w^\n) = 0 )
\nonumber\\
& \le \frac{1}{\zeta_L} \max_{w\in W}\| \nabla L_m(w) - \nabla L(w) \|_2,
\qquad \text{(see \eqref{eq:near_st})}
\end{align}
which completes the proof of Lemma \ref{lem:ws_to_wn}. 

\section{Experimental Setup Details}
\label{sec:exp_setup}
\subsection{Per-Iteration Computational Complexity}
\label{subsec:comp_complexity}

The gradient of the function
\begin{equation}
    h(z) = \dfrac{1}{2} \norm{AG(z) - b}_2^2
\end{equation}
follows the formula
\begin{equation}
    \grad h(z) = \grad G(z) A^\top (AG(z) - b)
\end{equation}
which involves one forward pass through the network $G$, in order to compute $G(z)$,
as well as one backward pass to compute $\grad G(z)$, and finally matrix-vector
products to compute the final result.

On the other hand our ADMM first computes the iterate $z_{t+1}$ with
gradient descent on
the augmented lagrangian \eqref{eq:defn_aug_lagr}
 as \begin{equation}
    z_{t+1} = z_t - \beta \grad_z \Lagr_\rho(w_t, z_t, \lambda_t) = - \grad G(z_t)
    \lambda_t^\top - \rho \nabla G(z_t)(w_t-G(z_t))^\top
\end{equation}
which involves one forward and one backward pass on the network $G$, as well as matrix-vector
products. Then we perform the exact minimization procedure on the $w$ variable, 
%, as
%described in \ref{subsec:fast_exact}, 
 which requires recomputing $G(z)$ on the new iterate
$z_{t+1}$, involving one forward pass through the network, as well as the matrix-vector
operations as described before. Recomputing the quantity $w_{t+1} - G(z_{t+1})$ is immediate
upon which the dual stepsize $\sigma_{t+1}$ can be computed at negligible cost. Finally the
dual variable update reads as
\begin{equation}
    \lambda_{t+1} = \lambda_t + \sigma (w_{t+1} - G(z_{t+1}))
\end{equation}
which involves only scalar products and vector additions of values already computed.
All in all each GD iteration involves one forward and one backward pass, while
ADMM computes two forward and one backward pass. Both algorithms require a few
additional matrix-vector operations of similar complexity. For networks with
multiple large layers, as usually encountered in practice, the complexity per
iteration can then be estimated as the number of forward and backward passes,
which are of similar complexity.

\subsection{Parameter Tuning}
\label{subsec:alg_parameters}
\begin{figure}[h]
    \centering
    \begin{minipage}{0.4\textwidth}
        \centering
        \includegraphics[scale=0.4]{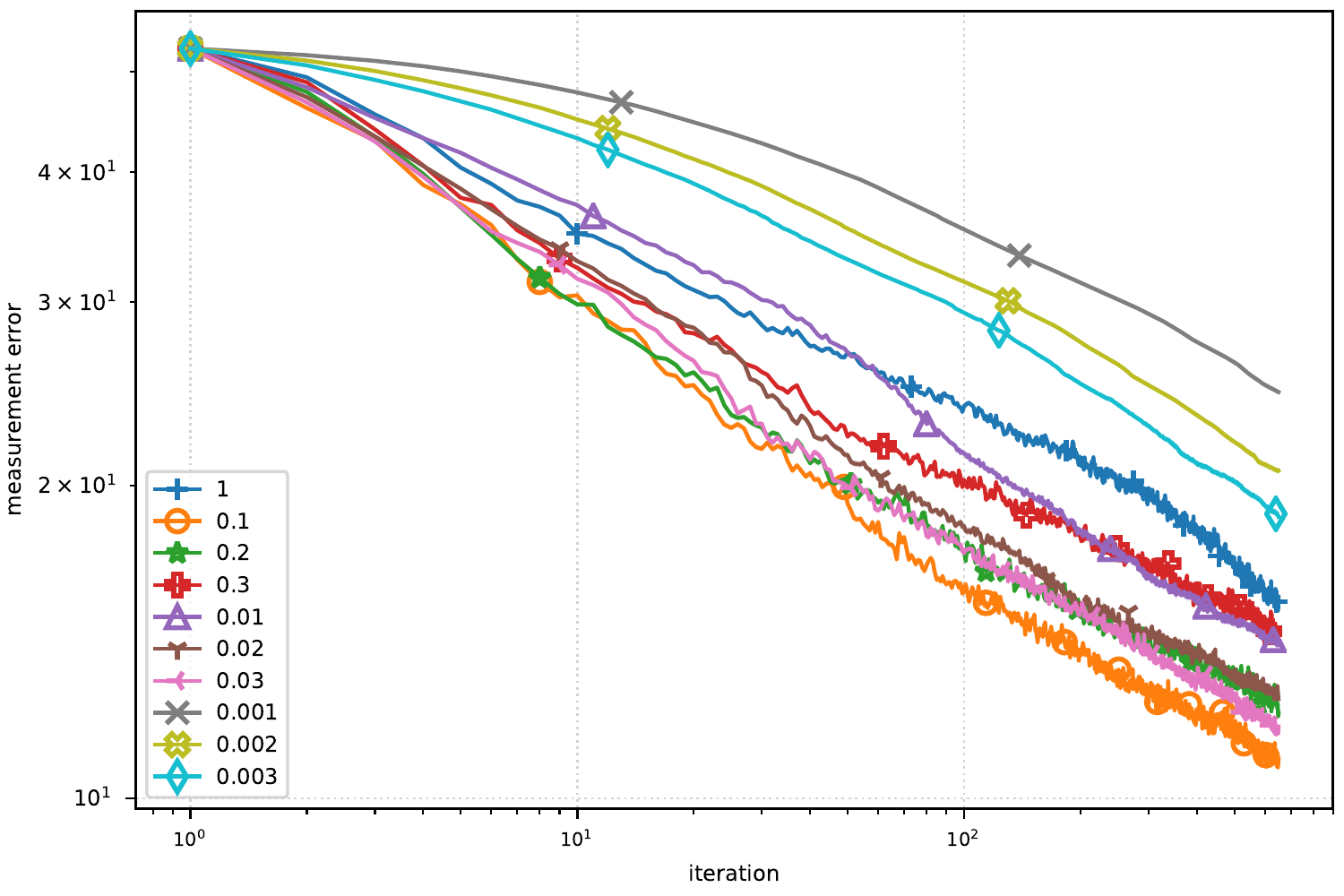}
    \end{minipage} \qquad
    \begin{minipage}{0.4\textwidth}
    \centering
        \includegraphics[scale=0.4]{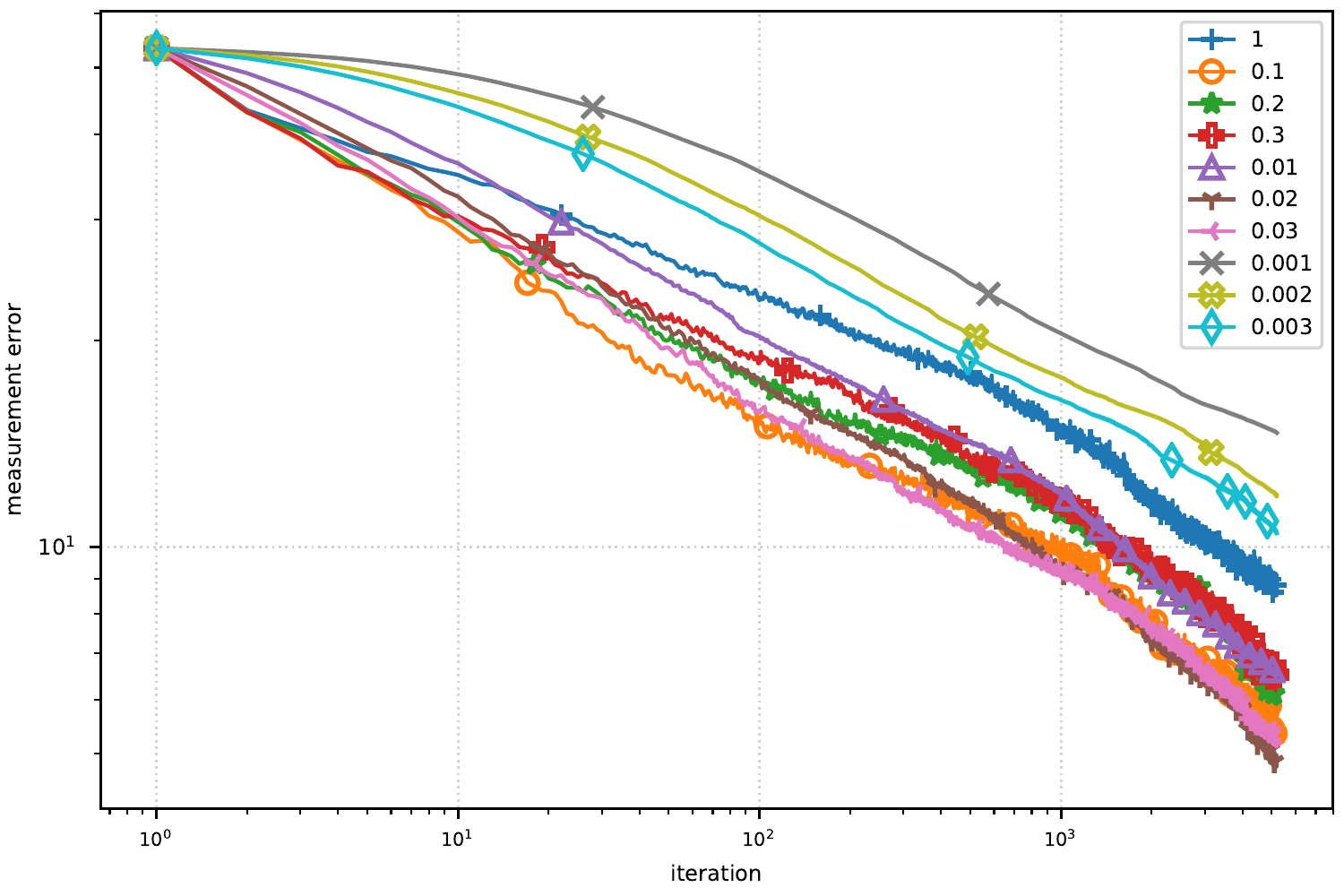}
    \end{minipage}
\caption{Performance of GD on the compressive sensing task for different
    step sizes. MNIST dataset. 156 (top) and 313 (bottom) linear measurements.}
\label{fig:gd_tuning_mnist}
\end{figure}

\begin{figure}[h]
    \centering
    \begin{minipage}{0.4\textwidth}
        \centering
        \includegraphics[scale=0.4]{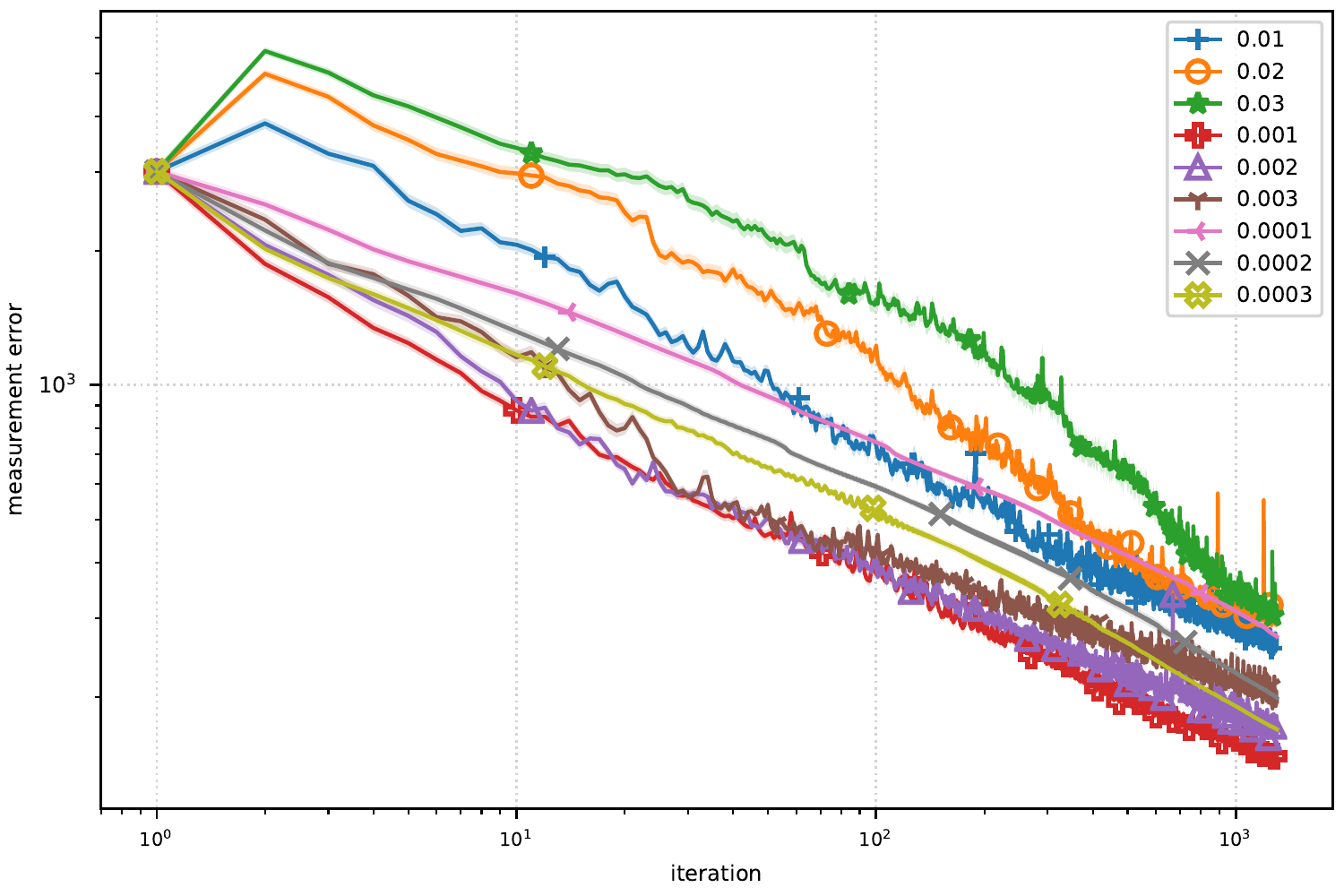}
    \end{minipage} \qquad
    \begin{minipage}{0.4\textwidth}
    \centering
        \includegraphics[scale=0.4]{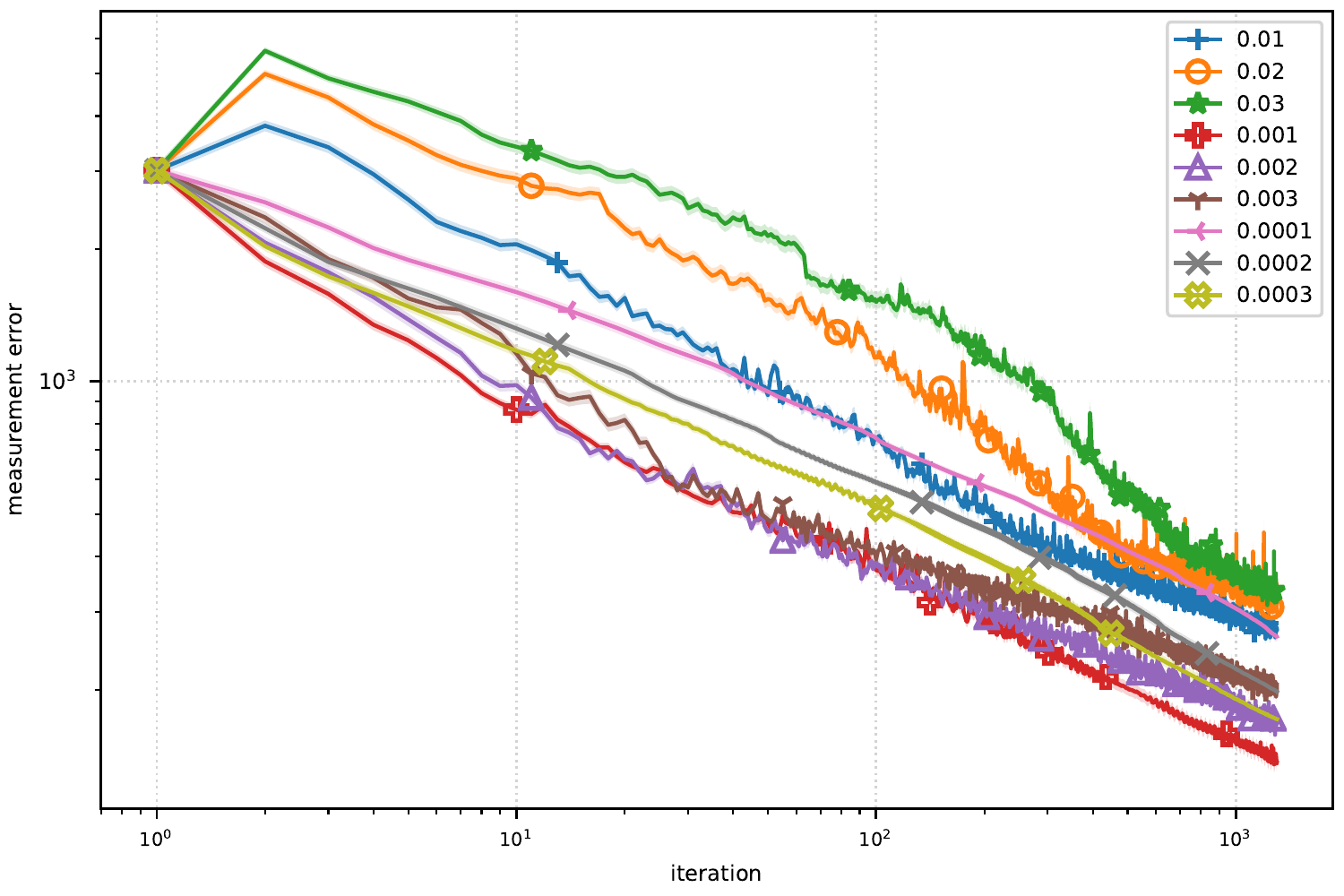}
    \end{minipage}
\caption{Performance of GD on the compressive sensing task for different
    step sizes. CelebA dataset. 2457 (top) and 4915 (bottom) linear measurements.}
\label{fig:gd_tuning_celeba}
\end{figure}

We run a grid search for the gradient descent (GD) algorithm
In order to do so we fix a number of
iterations and compare the average objective function over a batch of 100
random images and choose the best performing parameters. We repeat
the tuning in all possible escenarios in the experiments. The results
figures \ref{fig:gd_tuning_mnist} - \ref{fig:gd_tuning_celeba} (GD,
Compressive sensing setup).

\subsection{Fast Exact Augmented Lagrangian Minimization with Respect to Primal Variable $w$}
\label{subsec:fast_exact}
In the compressive sensing setup, the augmented lagrangian takes the form
\begin{equation}
    \label{eq:auglagr_appendix}
    \Lagr_\rho(w, z, \lambda) := \dfrac{1}{2} \norm{Aw-b}_2^2 + \langle \lambda, w-G(z)\rangle 
    +\dfrac{\rho}{2} \norm{w-G(z)}_2^2
\end{equation}
with respect to $w$, this is a strongly convex function which admits a unique minimizer
given by the first order optimality condition
\begin{equation}
    \grad_w \Lagr_\rho(w, z, \lambda) = A^\top(Aw-b)+\lambda + \rho(w-G(z))=0
\end{equation}
with solution
\begin{equation}
    w^* = (A^\top A + \rho I)^{-1}(-\lambda +G(z) +A^\top b)
\end{equation}
Given the SVD of $A=USV^\top $ we have $A^\top A=VDV^\top$, where $D$ corresponds to
the diagonal matrix with the eigenvalues of $A^TA$. We then have that
$A^\top A + \rho I=V(D + \rho I)V^\top$ so that
\begin{equation}
    w^* = V(D + \rho I)V^\top(-\lambda + G(z) + A^\top b)
\end{equation}
which involves only a fixed number of matrix-vector products per-iteration.

\end{document}